\documentclass{article} %
\usepackage[T1]{fontenc}
\newif\ifarxiv
\arxivtrue
\ifdefined\usearxiv \arxivtrue \fi
\ifarxiv
  \usepackage{arxiv}
  \usepackage{fnpos}
  \makeFNbelow
  \renewcommand{\And}{\par\medskip}
  \providecommand{\AND}{\par\medskip}
  
\else
  \usepackage{iclr2027_conference,times}
\fi

\usepackage{amsmath,amsfonts,bm}

\def\eqref#1{equation~\ref{#1}}

\def\1{\bm{1}}

\DeclareMathAlphabet{\mathsfit}{\encodingdefault}{\sfdefault}{m}{sl}
\SetMathAlphabet{\mathsfit}{bold}{\encodingdefault}{\sfdefault}{bx}{n}

\usepackage{float}
\floatstyle{ruled}
\newfloat{algorithm}{tbp}{loa}
\floatname{algorithm}{Algorithm}
\usepackage{hyperref}
\usepackage{url}
\usepackage{booktabs}
\usepackage{graphicx}
\usepackage{placeins}
\usepackage{wrapfig}
\usepackage{tikz}
\usepackage{afterpage}
\usepackage{enumitem}
\usetikzlibrary{decorations.pathmorphing, arrows.meta, positioning}
\usepackage[noend]{algpseudocode}
\usepackage{xspace}
\usepackage{enumitem}
\usepackage{cleveref}

\newcommand{\methodname}{Actor-Critic with Action Chunking\xspace}
\newcommand{\methodshort}{AC2\xspace}

\newcommand{\lukeappendixfig}[4]{%
\begin{figure}[H]
\centering
\includegraphics[width=#1]{luke_figures/appendix/#2.pdf}
\caption{#3}\label{#4}
\end{figure}}

\title{Trust the Critic More}

\ifarxiv
\author{Kaiyue Wen*, Luke Bailey*, Arvind Mahankali, Tengyu Ma}
\affiliation{Stanford University}
\else
\author{Antiquus S.~Hippocampus, Natalia Cerebro \& Amelie P. Amygdale \thanks{ Use footnote for providing further information
about author (webpage, alternative address)---\emph{not} for acknowledging
funding agencies.  Funding acknowledgements go at the end of the paper.} \\
Department of Computer Science\\
Cranberry-Lemon University\\
Pittsburgh, PA 15213, USA \\
\texttt{\{hippo,brain,jen\}@cs.cranberry-lemon.edu} \\
\And
Ji Q. Ren \& Yevgeny LeNet \\
Department of Computational Neuroscience \\
University of the Witwatersrand \\
Joburg, South Africa \\
\texttt{\{robot,net\}@wits.ac.za} \\
\AND
Coauthor \\
Affiliation \\
Address \\
\texttt{email}
}
\fi

\begin{document}
\raggedbottom

\maketitle
\ifarxiv
  {\renewcommand{\thefootnote}{}%
   \footnotetext{\hspace*{-1.8em}Correspondence to 
   \{kaiyuew,ljbailey\}@stanford.edu\hspace{4em}Code available at \url{https://github.com/WhenWen/AC2}\hspace{2em}
   * Equal contribution 
   }
   }%
\fi
\newcommand{\tm}[1]{{\color{blue}[TM: #1]}}

\begin{abstract}
Standard language model RL algorithms 
credit every token of a long rollout with
the same advantage determined by the terminal reward.
Actor--critic methods
can provide finer-grained credit assignment, but learned critics are generally
considered too inaccurate to trust 
when training LLMs with RL.
In recent works, even when a critic is present,
it is used only for baseline estimation, so every trajectory must be rolled out 
to its terminal reward \citep{venkatraman2026critique,pan2026evpo}.
We introduce \methodname (\methodshort) that removes the need to roll every trajectory to completion.
\methodshort instead assigns credit to action
chunks: short continuations of prefixes
of past trajectories. A
learned critic scores the state reached at the
end of each action chunk, allowing the policy
to update without observing a terminal reward. 
We make critic-based credit assignment reliable
through three design choices. First, we
introduce \emph{local readiness} which 
uses critic-based
updates on a problem only when the critic is
sufficiently accurate on that particular problem.
Second, when available, we provide the critic 
with a reference solution from a previous
successful rollout. Third, we assign credit
over action chunks of 10k tokens rather
than individual tokens, giving the critic a
more meaningful portion of the trajectory 
to evaluate.
We train Qwen3-4B on FineProofs-RL using
\methodshort and evaluate on IMO-ProofBench.
\methodshort exceeds GRPO's peak validation score of 18.5\% using $2.5\times$ 
fewer decoding FLOPs. This gain comes from
two sources, (1) \methodshort requires 25\% fewer
training steps to reach this score, and (2) each step
generates fewer tokens because the policy does
not need to continue every trajectory to completion. 
Conceptually, we demonstrate that we can 
remove the need 
to roll out every trajectory to completion,
opening up a large previously 
unexplored design space for LLM RL 
algorithms.
\end{abstract}

\definecolor{gruvyellow}{HTML}{D79921}
\definecolor{gruvgreen}{HTML}{98971A}
\definecolor{gruvblue}{HTML}{458588}
\newcommand{\colS}[1]{{\color{gruvyellow}#1}}
\newcommand{\colA}[1]{{\color{gruvgreen}#1}}
\newcommand{\colR}[1]{{\color{gruvblue}#1}}
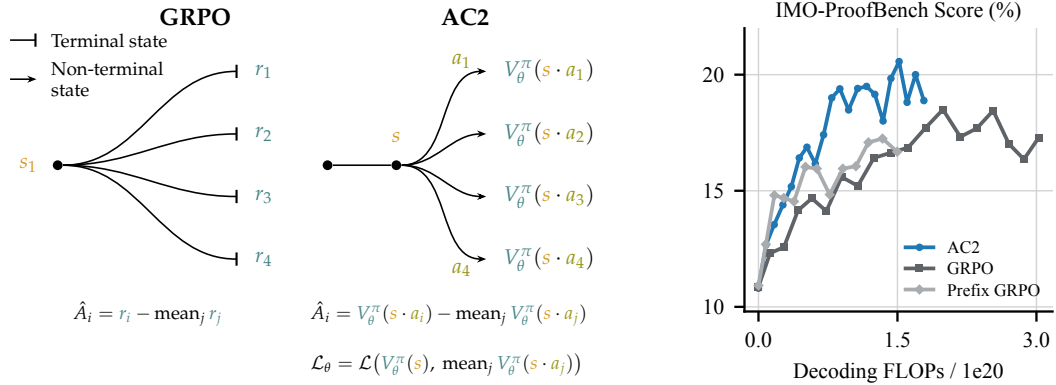
\begin{figure}[!b]
\centering
\begin{minipage}[c]{0.66\textwidth}
\centering
\begin{tikzpicture}[
  x=0.70cm, y=1.035cm,  %
  traj/.style={line width=0.6pt},
  term/.style={traj, -{Bar[width=5pt, line width=0.8pt]}},
  cont/.style={traj, -{Stealth[length=3.5pt]}},
  root/.style={circle, fill=black, inner sep=1.3pt},
  lab/.style={font=\footnotesize},
  formula/.style={font=\scriptsize},
]
\begin{scope}[shift={(-0.3,0)}]
  \draw[term] (0, 1.6) -- (0.5, 1.6);
  \node[font=\scriptsize, right=2pt] at (0.5, 1.6) {Terminal state};
  \draw[cont] (0, 1.1) -- (0.5, 1.1);
  \node[font=\scriptsize, right=2pt, align=left] at (0.5, 1.1) {Non-terminal\\state};
\end{scope}
\begin{scope}[shift={(0.6,0)}]
  \node[font=\bfseries\small] at (2.6, 1.9) {GRPO};
  \node[root] (g0) at (0, 0) {};
  \node[lab, left=2pt of g0] {$\colS{s_1}$};
  \foreach \i/\y in {1/1.2, 2/0.4, 3/-0.4, 4/-1.2} {
    \draw[term] (g0) to[out=0, in=180] (3.4, \y);
    \node[lab, right=3pt] at (3.4, \y) {$\colR{r_{\i}}$};
  }
  \node[formula] at (1.7, -1.9) {$\hat{A}_i = \colR{r_i} - \mathrm{mean}_j\,\colR{r_j}$};
\end{scope}
\begin{scope}[shift={(5.7,0)}]
  \node[font=\bfseries\small] at (2.6, 1.9) {\methodshort};
  \node[root] (a0) at (0, 0) {};
  \node[root] (ap) at (1.3, 0) {};
  \draw[traj] (a0) -- (ap);
  \node[lab, above=3pt of ap] {$\colS{s}$};
  \foreach \i/\y/\l/\pos/\off in {1/1.2/a_1/above/2pt, 2/0.4//above/1pt, 3/-0.4//below/1pt, 4/-1.2/a_4/below/2pt} {
    \draw[cont] (ap) to[out=0, in=180]
      node[lab, pos=0.78, \pos=\off] {\ifx\l\empty\else$\colA{\l}$\fi} (3.0, \y);
    \node[lab, right=3pt] at (3.0, \y) {$\colR{V^\pi_\theta}(\colS{s} \cdot \colA{a_{\i}})$};
  }
  \node[formula, anchor=west] at (-0.5, -1.9) {$\hat{A}_i = \colR{V^\pi_\theta}(\colS{s} \cdot \colA{a_i}) - \mathrm{mean}_j\,\colR{V^\pi_\theta}(\colS{s} \cdot \colA{a_j})$};
  \node[formula, anchor=west] at (-0.5, -2.55) {$\mathcal{L}_\theta = \mathcal{L}\big(\colR{V^\pi_\theta}(\colS{s}),\ \mathrm{mean}_j\,\colR{V^\pi_\theta}(\colS{s} \cdot \colA{a_j})\big)$};
\end{scope}
\end{tikzpicture}
\end{minipage}\hfill
\begin{minipage}[c]{0.32\textwidth}
\centering
\includegraphics[width=\linewidth]{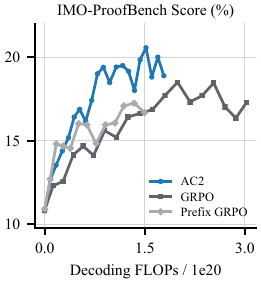}
\end{minipage}
\caption{Advantage estimation in GRPO and \methodshort. GRPO rolls every sample
out from the root to a terminal reward. \methodshort starts from a replayed
prefix $s$, generates a short chunk $a_i$ per sample, and reads the critic at the
chunk's end, $V^\pi_\theta(s \cdot a_i)$, where $\cdot$ denotes concatenation. The same group
also trains the critic. $\mathcal{L}$ is a classification loss that fits the
critic's prediction at the prefix to the group-mean value, its
bootstrapped target (\cref{sec:updates}). Right: mean score against decoding
compute for \methodshort, GRPO, and Prefix GRPO, which applies GRPO to full-length
continuations of replayed prefixes (\cref{sec:main-results}).}
\label{fig:method-overview}
\end{figure}

\section{Introduction}
\label{sec:intro}

Standard reinforcement learning algorithms today, such as Group Relative Policy
Optimization (GRPO)~\citep{shao2024deepseekmath}, treat each
sampled rollout
as a single action. As a result, every token
of a rollout is credited
with the same advantage
that depends solely on the terminal reward. This provides only coarse
credit assignment. For example, a minor calculation error at the end 
of a math proof can cause an otherwise
correct and valuable reasoning trajectory to be penalized.

If we instead view a rollout as a sequence of many small actions, actor--critic
methods can assign credit to each action individually, providing dense learning
signal throughout the trajectory \citep{haarnoja2018soft,konda1999actor}. 
However, current LLM reinforcement learning methods
largely avoid using critics for fine-grained credit assignment, and at most use
them as baselines while still relying on terminal rewards for advanatage estimation \citep{pan2026evpo,venkatraman2026critique}.
In essence, learned critics are considered too inaccurate to 
be used in isolation without any terminal reward~\citep{kazemnejad2024vineppo}.

We revisit this reliance on terminal rewards by introducing \methodname (\methodshort), which instead assigns credit to \emph{action chunks}: short continuations of prefixes of past trajectories. A learned critic scores the state reached at the end of each action chunk, allowing the policy to update without rolling 
out to a terminal reward (\Cref{fig:method-overview}). Formally, let $\pi_\theta$ and $V^\pi_\theta$ be the policy and value function. From a partial chain-of-thought prefix $s$, we sample $g$ 
token chunks
$a_i$ of the same length.
Each continuation is scored by the critic at its endpoint, $v_i=V^\pi_\theta(s\cdot a_i)$, 
where $\cdot$ denotes concatenation, and receives advantage $\hat A_i=v_i-1/g\sum_{j}v_j$.
No terminal reward is required for these updates.

Removing the need for a terminal reward on 
every update opens up a 
part of the design space for LLM 
reinforcement learning that has not
been explored.
\methodshort benefits from this in
two ways. Firstly, 
each step can generate far fewer 
tokens by only rolling out 
groups of action-chunks instead 
of full trajectories. Secondly, 
the actor receives finer-grained 
supervision of a single 
advantage for each action-chunk as opposed 
to trajectory.

We identify three components of the algorithm that make fully critic based advantage estimates
reliable.
First, we introduce \emph{local readiness}, which measures the critic's error on the particular problem the actor is rolling out. Before the critic becomes sufficiently accurate on a problem, we generate a group of continuations to completion from a sampled prefix. We use their terminal rewards as endpoint values for policy updates and fit the critic to the average terminal reward. After the problem meets the readiness criteria, we instead generate action chunks, use the critic to assign endpoint values for policy updates, and update the critic using the average predicted value.
Second, when available, we provide reference solutions from previous 
successful rollouts on the problem to the critic 
in context.
Third, 
we use an action-chunk budget of 10k tokens with a full rollout budget of 50k
tokens, providing a more meaningful unit of credit assignment than a single token.

We train Qwen3-4B-Thinking-2507 \citep{yang2025qwen3} on FineProofs-RL~\citep{lmprovers2026qednano}
and evaluate on IMO-ProofBench~\citep{luong2025robust}, a collection 
of math olympiad level problems.
\methodshort exceeds GRPO's peak mean score of
18.50\% using $2.5\times$ fewer decoding FLOPs 
(\Cref{fig:method-overview}). This gain comes
from two places: (1) \methodshort 
requires 25\% fewer training steps,
surpassing GRPO's peak in 90 as opposed 
to 120 steps, and 
(2) \methodshort generates fewer tokens per update because it does not need to
continue every trajectory to completion.
\methodshort also achieves a higher peak mean score of 20.57\%, compared with
18.50\% for GRPO.
Decoding FLOPs are highly correlated with GPU hours in our main RL run
(\Cref{app:gpu-hours}). We use them as a normalized compute proxy
throughout the paper, since different experiments use different numbers of GPUs.

We provide evidence that all three components, local readiness, 
providing reference solution to critic, and action-chunking, are all integral to the success of 
the agorithm. We find that runs without local 
readiness plateau early because they apply 
an imprecise critic to some problems
(\Cref{fig:component-ablations} right).
When we reduce action-chunk size from 10k to 2k tokens, 
performance plateaus earlier than the 10k run 
(\Cref{fig:component-ablations} right).
Finally, we show the critic error is lower when provided 
with a correct reference solution in context
(\Cref{fig:main-results}, fourth panel).
We also demonstrate two variants of the AC2 algorithm 
that perform comparably, the first samples a single 
action-chunk for each critic-ready problem, 
using $\hat{A} = V^\pi_\theta(s\cdot c) - V_\theta^\pi(s)$.
The second 
uses a highly stale replay buffer for sampling 
trajectory prefixes
(\Cref{fig:component-ablations} left). We present these variants because they broaden the space of RL algorithms, for example, by enabling algorithms that leverage existing trajectories more extensively.

 Conceptually, our results 
 can be summarized as 
 showing that 
 LLM RL can
trust a learned critic
far more than current methods do. Once the critic is accurate, 
rolling every trajectory out to a terminal reward is no longer required, 
and removing that requirement opens up the design space. 
\methodshort is one example of what this makes
possible. 
It assigns credit to individual action
chunks, generates a small number 
of tokens per step, and learns 
from off-policy initial states drawn 
from a replay buffer.

\section{Preliminaries}
\label{sec:prelim}

We consider the problem 
of reinforcement learning 
with verifiable rewards.
Let $\mathcal D$ be the question distribution and $\mathcal V$ the model vocabulary.
A policy $\pi_\theta$ with parameters $\theta$ maps a question $x \sim \mathcal{D}$ to an autoregressively generated
response of length $T$ tokens, $y = (y_1, \dots, y_T)$, $y_t \in \mathcal{V}$, which a deterministic verifier scores
$R(x, y) \in [0, 1]$. The learning objective is the expected verifier score
$J(\theta) \;=\; \mathbb{E}_{x \sim \mathcal{D}} \, \mathbb{E}_{y \sim \pi_\theta(\cdot \mid x)}
  \big[\, R(x, y) \,\big]$.
$R$ is defined only on complete responses.

\textbf{Token-level MDP.}
We will consider the following MDP. Let $s_t = x \cdot  y_{<t}$ where $\cdot$ denotes string concatenation,
with $s_1 = x$ and $a_t = y_t \in \mathcal{V}$.
The transition dynamics are simply $s_{t+1} = s_t \cdot a_t$. The reward of the MDP is 
$r_t = 0$ for $t < T$ and $r_T = R(x,y)$. 
Let $V^\pi(s)$ denote the expected terminal
verifier score when following $\pi$ from prefix $s$
(equal to $R(x,y)$ for a complete response) and $Q^\pi(s,a)$ the expected score
after taking $a$ at $s$ and following $\pi$.
Since transitions are deterministic and rewards are terminal-only we have 

\begin{equation}
  \label{eq:q-collapse}
  \begin{aligned}
    Q^\pi(s, a) \;=\; V^\pi(s \cdot a),  \quad
    A^\pi(s, a) \;=\; Q^\pi(s,a) - V^\pi(s) \;=\; V^\pi(s \cdot a) - V^\pi(s).
  \end{aligned}
\end{equation}

One can also reform the MDP to have $a$ denote a fixed-size chunk of consecutive 
tokens, and these identities are unchanged.

\textbf{Policy gradients and advantage estimation.}
The policy gradient theorem \citep{sutton1999policy} states that, for any action-independent
baseline $b(s)$ and policy $\pi_\theta$ with state visitation distribution $d^\pi$,
\begin{equation}
  \label{eq:pg}
  \nabla_\theta J(\theta) \;\propto\; \mathbb{E}_{s \sim d^{\pi}, \, a \sim \pi_\theta}
  \Big[\, \nabla_\theta \log \pi_\theta(a \mid s) \,\big( Q^\pi(s,a) - b(s) \big) \Big].
\end{equation}
Note that the reward to go does not appear in
\Cref{eq:pg}. An algorithm needs some estimate $\hat A_t$ of $A^\pi(s_t,a_t)$, and the choice of
estimator is what separates existing LLM policy gradient methods. 
Let $V$ be an approximate state-value function, set to $R(x,y)$ on a complete response,
and let $\lambda\in[0,1]$ control
the decay of residual weights in Generalized Advantage Estimation (GAE)
\citep{schulman2018highdimensionalcontinuouscontrolusing}.
With the one-step residual $\delta_t^V = V(s_{t+1}) - V(s_t)$,
\begin{equation}
  \label{eq:gae}
  \hat A_t^{\mathrm{GAE}(\lambda)} \;=\; \sum_{l=0}^{T-t} \lambda^l \, \delta_{t+l}^V.
\end{equation}
In the case when $V \neq V^\pi$, for example in practice when we use 
a learnt value model, low $\lambda$ corresponds to low variance and high bias estimators, 
and high $\lambda$ the converse. 
Importantly, once $\lambda > 0$, estimating $\hat A_t^{\mathrm{GAE}(\lambda)}$ requiers evaluating the terminal reward. The two $\lambda$ endpoints take particularly simple forms:

\begin{equation}
  \label{eq:lambda1}
  \hat A_t^{\mathrm{GAE}(0)} = V(s_t \cdot a_t) - V(s_t),
  \qquad\qquad
  \hat A_t^{\mathrm{GAE}(1)} = R(x,y) - V(s_t).
\end{equation}

\noindent
The first is the single residual $\delta_t^V$, the second follows by telescoping
the residuals through the complete response.
At $\lambda = 0$, advantages at nonterminal transitions are estimated
entirely using the learned value function, without the terminal reward.
At $\lambda = 1$ it is the realized verifier score, with the value function serving
\emph{only as a baseline}.

\textbf{Existing RLVR methods estimate the advantage at $\lambda \approx 1$.}
Modern RLVR pipelines optimize $J(\theta)$ with a PPO-style clipped surrogate
\citep{schulman2017ppo}. Let $\theta_{\mathrm{old}}$ be the fixed parameters of the policy
that generated the sampled responses and
$\rho_{t}(\theta) = \pi_\theta(a_t \mid s_t) /\pi_{\theta_{\mathrm{old}}}(a_t \mid s_t)$ denote the off-policy importance sampling ratio.
With lower and upper clipping parameters $\epsilon_{\mathrm{low}}$ and $\epsilon_{\mathrm{high}}$, the PPO loss is 
\begin{equation}
  \label{eq:ppo}
  \mathcal{L}(\theta) \;=\; -\,\mathbb{E}_{x \sim \mathcal D,\; y \sim \pi_{\theta_{\mathrm{old}}}(\cdot \mid x)}\Big[ \textstyle\sum_t
  \min\big( \rho_t(\theta)\, \hat A_t, \;
  \mathrm{clip}(\rho_t(\theta), 1-\epsilon_{\mathrm{low}}, 1+\epsilon_{\mathrm{high}})\, \hat A_t \big) \Big].
\end{equation}
Most RLVR methods differ in the
estimator $\hat A_t$. All of them use $\lambda$ (\Cref{eq:gae}) very close to $1$.
For comparison, PPO-based RLHF \citep{ouyang2022instructgpt}
trains a token-level value network $V_\theta$ and
uses $\hat A_t$ from \Cref{eq:gae} with $\lambda \approx 0.95$. 
In RLVR, GRPO \citep{shao2024deepseekmath} replaces the learned value network
with a baseline computed from a group of sampled responses.
For each question $x$, it samples $g$ responses
$y_1, \dots, y_G \sim \pi_{\theta_{\text{old}}}(\cdot \mid x)$
and computes their mean reward, $V = 1/g\sum_{j=1}^g R(x,y_j)$.
Every token in response $y_i$ receives the same advantage,
$\hat A_{i,t} = R(x,y_i) - V$, optionally divided by the group's reward standard deviation.
The group mean is a Monte Carlo estimate of the root-state value $V^\pi(s_1)$,
so GRPO uses the $\lambda = 1$ estimator with a shared baseline.

Other RLVR methods learn a value model but use its predictions only as a baseline,
retaining the $\lambda = 1$ advantage estimator.
Like GRPO, these methods estimate advantages from the realized terminal reward but 
they differ in how they construct the baseline
(more details in \Cref{sec:related-works}).
 Our method differs from the previous works by choosing $\lambda = 0$, using the critic to remove the dependency on terminal rewards for some problems.

\section{\methodname}
\label{sec:method}

We present \methodname (\methodshort). 
Let $\mathcal{D}$
be a set of training problems
that admit a terminal reward.
\methodshort trains a policy $\pi_\theta$ 
and value model $V^\pi_\theta$ that 
share parameters $\theta$. Our method 
is motivated by a single guiding principle. 
\emph{When we know the value function is accurate, 
we use it to update the policy}. 
Each step of the \methodshort proceeds 
as follows (see \Cref{alg:ac2}):

\begin{enumerate}[leftmargin=*]
	\item \textbf{Problem sampling.} We maintain 
	a replay buffer $\mathcal{B}$ of complete rollouts  
	on prior problems. At each step we sample 
	$n_{\text{refill}}$ new problems from $\mathcal{D}$ 
	and $n_{\text{batch}}$ 
	complete trajectories from $\mathcal{B}$. 
	\item \textbf{Policy sampling.}
	For each sampled new problem, we do a rollout to the end, 
	and use these to refill the buffer for the next step. 
	For each buffer sampled trajectory, we cut 
	to a random prefix of tokens $s$, 
	and rollout a group number $g$ of 
    token blocks from 
	there. If our value model is deemed accurate on the 
	problem, which we call \emph{ready} (\Cref{sec:sampling}), then we rollout a chunk 
    of $b$
	tokens. If not then we rollout until the end 
	of the trajectory. Let $c_i$ denote 
	these sampled tokens for the $i$th entry 
	in the group. This gives us a group of 
	visited states $\{ s \cdot c_i \}_{i=1}^g$
	\item \textbf{Advantage estimation.} 
	To calculate the advantage 
	we need the expected return from each continuations 
	state $s \cdot c_i$. If $s \cdot c_i$ is a complete trajectory this 
	is simply the terminal reward $r_i$.
	Otherwise we use the critic's 
	estimate $V^\pi_\theta(s \cdot c_i)$. Writing 
	$v_i$ for endpoint value, the advantage is
	$\hat{A}_i = v_i - \text{mean}_j(v_j)$. 
	\item \textbf{Actor update.} We take a clipped policy gradient 
	step with $\hat{A}_i$ on the newly sampled tokens, leaving the 
	replayed prefix $s$ as conditioning context that receives no loss.
	\item \textbf{Critic update.} We fit $V^\pi_\theta$ to each prefix's group-mean value $\text{mean}_i(v_i)$. If $V^\pi_\theta$ was
		not ready on the problem, then this target will 
		be the average terminal reward over the group. 
		If it was ready, the target will be an average over the 
		value model's own estimates $\text{mean}_j\big(V^\pi_\theta
		(s \cdot c_j)\big)$.
\end{enumerate}

\subsection{Problem and policy sampling}
\label{sec:sampling}

We begin by sampling states to generate actions for.
We keep a replay buffer $\mathcal{B}$ of complete 
rollouts (that are not necessarily correct) on 
prior problems. 
We sample $n_{\text{refill}}$ fresh problems from
$\mathcal{D}$ without replacement and $n_{\text{batch}}$ stored trajectories
from $\mathcal{B}$. For each of these $n_{\text{batch}}$ 
trajectories we choose a random cut position after the problem 
statement. We keep all tokens 
before the cut. We refer to this state as $s$.
Let $\mathcal{S}_{\text{refill}}$ be the set of $n_{\text{refill}}$ fresh
states, each a problem statement with an empty response, and let
$\mathcal{S}_{\text{batch}}$ be the set of $n_{\text{batch}}$ cut states.
For each problem in $\mathcal{S}_{\text{refill}}$, we do a single full 
rollout to the end, and add these to $\mathcal{B}$, which is 
simply a first-in-first-out queue with fixed 
capacity. We only use these rollouts to update the buffer, and they 
are not part of the actor and critic loss.

We use the trajectory prefixes in $\mathcal{S}_{\text{batch}}$ 
to update the actor and critic. We partition the problems into ready and unready sets
according to whether they have satisfied the readiness criteria, and write
$\mathcal{D}_{\text{ready}} \subseteq \mathcal{D}$ for the ready set.
A prefix is ready when its problem is. By default, 
a problem is not ready. Assume 
we have access to the error of the critic $\varepsilon$
on the problem from prior steps (we explain 
how this error is calculated in the following 
paragraph). 
A problem is deemed ready when three conditions hold: (1) the mean of $\varepsilon$
over every prefix sampled in the last five steps falls 
below a global threshold
$\tau_{\text{global}}$, (2) the error recorded 
for that problem from the prior step it was sampled 
falls below a tighter local threshold
$\tau_{\text{local}}$, and (3) at least one correct 
trajectory has been found by the policy on this problem.

For each prefix $s \in \mathcal{S}_{\text{batch}}$ we then sample a group of $g$
continuations $c_i \sim \pi_\theta(\cdot \mid s)$ from the
policy. If the problem is not ready, each continuation runs until the response
terminates and receives a terminal reward $r_i$. 
In this case we use the terminal reword to calculate 
the critic error $\varepsilon = \big|V^\pi_\theta(s) - \text{mean}_i(r_i)\big|$ and store it for future 
readiness calculations.
If the problem is ready, each continuation is a
single chunk of at most $b$ new tokens, after which generation stops.  
For a fraction $\alpha$ of ready problems (normally $\alpha = 1/4$)
we rollout full trajectories, so 
that terminal rewards continue to be recorded for them and
$\varepsilon$ remains measurable after a problem becomes ready. This 
is useful for (1) measuring the accuracy of the critic during the run 
and (2) for updating the critic on ready problems (see \Cref{sec:updates}).
We call this feature \emph{auditing}, and it  can be turned off by setting $\alpha = 0$
(see \Cref{sec:component-ablations} for this ablation).

Overall this gives us, for every $s \in \mathcal{S}_{\text{batch}}$, a group of
$g$ continuations $\{c_i\}_{i=1}^{g}$ and hence $g$ new states
$\{s \cdot c_i\}_{i=1}^{g}$. We write
$\mathcal{G} = \big\{(s, \{c_i\}_{i=1}^{g}) : s \in \mathcal{S}_{\text{batch}}\big\}$
for the set of prefix--group pairs. We use $\mathcal{G}$ to compute 
actor and critic updates.

\begin{algorithm}[t]
\caption{\methodname\ (\methodshort)}
\label{alg:ac2}
\begin{algorithmic}[1]
\Require problems $\mathcal{D}$, actor $\pi_\theta$
  and critic $V^\pi_\theta$, group size $g$, chunk length $b$
\State $\mathcal{B} \leftarrow \emptyset$, $\mathcal{D}_{\text{ready}} \leftarrow \emptyset$
\For{iteration $k = 1, 2, \dots$}
  \State Sample $\mathcal{S}_{\text{refill}} \subseteq \mathcal{D}$; cut
    $\mathcal{S}_{\text{batch}}$ from rollouts in $\mathcal{B}$
  \For{each $x \in \mathcal{S}_{\text{refill}}$}
    \State Generate one full response, judge it, insert into $\mathcal{B}$
  \EndFor
  \For{each $s \in \mathcal{S}_{\text{batch}}$}
    \State Sample $c_1, \dots, c_g \sim \pi_\theta(\cdot \mid s)$, to termination if
      the problem $\notin \mathcal{D}_{\text{ready}}$,
    \Statex \hspace{4.5em} else for at most $b$ tokens
    \State $v_i \leftarrow V^\pi_\theta(s \cdot c_i)$ if $s \cdot c_i$ is not
      complete, else $r_i$, for $i = 1, \dots, g$
    \State $\hat{A}_i \leftarrow v_i - (\sum_j v_j)/g$
    \State $\varepsilon \leftarrow \big|V^\pi_\theta(s) - (\sum_j v_j )/g\big|$
  \EndFor
  \State Add to $\mathcal{D}_{\text{ready}}$ each problem satisfying both
    readiness criteria based on $\varepsilon(s)$ (\Cref{sec:sampling})
  \State Update actor (\Cref{eq:ppo}) using $\hat A_i$ and critic
\EndFor
\end{algorithmic}
\end{algorithm}

\subsection{Advantage estimation}
\label{sec:advantage}

We use every pair $(s, \{c_i\}) \in \mathcal{G}$ to 
update the actor and critic. 
For each pair we wish to compute an advantage $\hat{A}_i$ for continuation
$c_i$.
By \Cref{eq:q-collapse}, the advantage of a block is
$V^\pi(s \cdot c_i) - V^\pi(s)$. We first consider the expected 
return $V^\pi(s \cdot c_i)$ for each new state $s \cdot c_i$. 
If $s \cdot c_i$ is a complete trajectory we know this exactly, 
is it the terminal reward $r_i$. If the trajectory is not 
complete, which is usually the case on critic ready problems 
that we sample a single action chunk for, we 
we use the critic, $V^\pi_\theta(s \cdot c_i)$.
Write $v_i$ for this endpoint value at $s \cdot c_i$, whether it is the
terminal reward or the critic's estimate. The advantage also needs $V^\pi(s)$, the
expected return from the prefix itself. 
Since
$V^\pi(s) = \mathbb{E}_{c \sim \pi}\big[V^\pi(s \cdot c)\big]$ the group-mean value $\text{mean}_j(v_j)$ is an
unbiased Monte-Carlo
estimate of it. 
Together these give
$\hat{A}_i \;=\; v_i - 1/g\sum_{j=1}^{g} v_j$.

Note that on critic ready problems, this is the 
the $\lambda = 0$ form of the GAE advantage in \Cref{eq:lambda1}, with the
baseline estimated from the group. 
On an unready problem, every $v_j$ is a terminal reward, 
and thus the advantage is the $\lambda = 1$ form.
Readiness can thus be interpretted as a per-problem switch 
to the $\lambda$ parameter of the advantage estimate. 
With $\lambda = 0$ we \emph{no longer need to observe the terminal reward},
and instead can update on a single action-chunk.

The critic $V^\pi_\theta$ share weights with the policy network. We query the critic through a prompt rather than a separate head. The prompt contains the
problem, the partial response, and, when the problem has previously been solved, a
known-correct solution. The model answers with a value 
in $\{0, 0.1, \dots, 1\}$, which we decode greedily and parse.  The exact instruction is given in
\Cref{app:implementation}.

\subsection{Actor and critic update}
\label{sec:updates}

The actor optimizes objective of
\Cref{eq:ppo} on the newly generated tokens,
holding $\hat{A}_i$ fixed across the tokens of $c_i$. 
Replayed prefix tokens are
conditioning context and receive no loss. 
We adapt $\epsilon_\text{high}$ based on the policy 
entropy following \citet{mai_thinking_1}.

The critic's target for a prefix is the group-mean value $\text{mean}_j(v_j)$, rounded to the
nearest grid value. Since the critic answers with a value on a discrete grid, we fit
it is as a next-token-prediction task. When the problem has a reference solution we train
the same target under both prompts, with and without the reference, at equal weight.
When there is no reference solution we train only the plain prompt. Prefix--target pairs are kept in a
first-in-first-out buffer of their own and each
critic step trains on a batch sampled from it, so a target is reused across
several updates before it ages out.

We optimize the same parameters $\theta$ for both objectives, using 
different optimizer state and learning rate for each. For all our experiments, at each step we update the 
actor with two steps. We 
interleave the critic update between the two. 
We defer details of this algorithm to~\Cref{app:implementation}.

\section{Experiments}
\label{sec:results}

We test AC2 by
training Qwen3-4B-Thinking-2507 on FineProofs-RL~\citep{lmprovers2026qednano}, a set of roughly 5,200
Olympiad proof problems from international competitions and AoPS.
We evaluate on IMO-ProofBench~\citep{luong2025robust}, 60 Olympiad problems
curated with grading rubrics, and reference solutions.
We use 
DeepSeek-V4-Flash~\citep{deepseekai2026deepseekv4highlyefficientmilliontoken} to judge proof correctness,
scoring each generated proof from 0 to 7. The judge 
is prompted with the correct reference solution and problem-specific 
rubric as well as the candidate proof to grade for validation. We do not provide the reference solution or the problem-specific rubric to the judge during training.

\textbf{Methods.}
We compare AC2 with GRPO and Prefix GRPO, a GRPO variant that 
starts its rollouts from replayed prefixes.
AC2 uses group size $g=16$, chunk length $b=10{,}000$, and auditing with $\alpha = 1/4$,
and empty initial replay buffers.
Prefix GRPO uses the same actor trajectory buffer system as AC2, but 
rolls out full-length continuations of prefixes 
and uses terminal rewards for advantage estimation identical 
to GRPO.
We select the GRPO learning rate by peak validation mean score
(\Cref{app:grpo-tuning}).
See \Cref{tab:training-settings} for 
hyperparameters and \Cref{app:implementation}
for additional details).

\textbf{Compute accounting.}
We track policy performance against a measure of compute used. 
The standard measure would be number of steps. This 
is imperfect in our setting as AC2, Prefix GRPO, and GRPO 
use different amounts of compute, and accordingly wall-clock 
time for each step. Specifically, GRPO uses the most compute 
as it requires rolling out all trajectories to completion.
A strong candidate to use is GPU hours 
(as used by \citet{khatri2026art}), however we conduct our 
experiments using different numbers of GPUs, 
and due to differing communication overhead GPU hours are not 
directly comparable between runs. Instead we find Decoding 
FLOPs (the FLOPs used in policy rollouts) correlate strongly with GPU hours in our main run 
(\Cref{app:compute}) and so use this hardware-agnostic measure instead. We also report performance against steps for full clarity.

\subsection{Main Results}
\label{sec:main-results}

\textbf{AC2 is more compute efficient and data efficient than GRPO.} We 
present results in \Cref{fig:method-overview}. 
AC2 reaches higher validation score with fewer steps and less decoding compute
than GRPO (\Cref{fig:method-overview}).
AC2 first exceeds GRPO's peak mean score of 18.50\% at
0.79e20 FLOPs compared to
1.99e20 for GRPO. This 
is a $2.5\times$ compute efficiency gain over GRPO.
Here both 
algorithms are also acting under a data constrained setting 
as there is a fixed number of training problems that 
we epoch. Under this constraint, AC2 reaches 
a higher peak score of 
20.57\% over 18.5\% for GRPO.

\begin{figure}[t]
\centering
\includegraphics[width=0.75\textwidth]{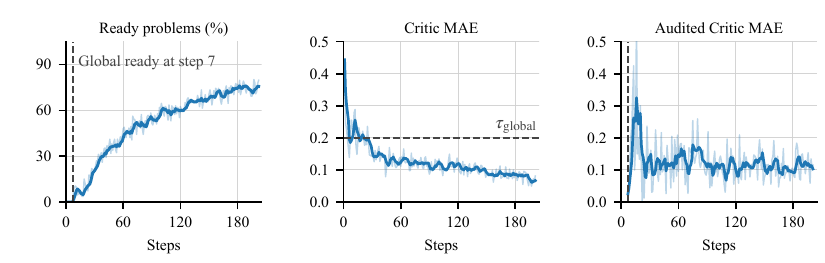}%
\includegraphics[width=0.25\textwidth]{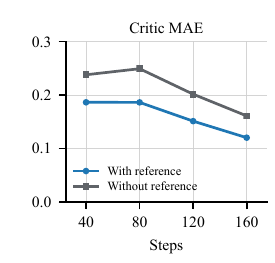}
\caption{\textbf{Critic readiness and accuracy for \methodshort.}
The first three panels show critic metrics over the course of the main \methodshort run, shown as the
\textcolor[HTML]{1F77B4}{blue} line in \Cref{fig:method-overview}.
First: fraction of sampled problems that are ready.
Second: critic error over all sampled prefixes,
$\big|V^\pi_\theta(s) - \text{mean}_j(v_j)\big|$, where $v_j$ is the terminal
reward or the critic's value of the continuation.
Third: critic error on audited groups of critic ready problems,
$\big|V^\pi_\theta(s) - \text{mean}_j(r_j)\big|$, where every continuation runs to
completion, so the target is a mean of terminal rewards only.
In the first three panels, faint lines show per-step values, dark lines average these over the trailing five steps.
Fourth: critic MAE with and without the reference solution in its prompt.
}
\label{fig:main-results}
\end{figure}

\textbf{The replay-buffer state distribution alone does not explain AC2's gains.}
We test whether AC2's gains come from the state distribution induced by replay
using Prefix GRPO. Prefix GRPO falls well behind AC2 when measured by Decoding FLOPs
(\Cref{fig:method-overview}).

\textbf{The critic becomes trustworthy early and stays so.}
The first three panels of \Cref{fig:main-results} track the two quantities that govern how much
\methodshort relies on its critic. The global readiness threshold is first
crossed at the end of step 7, after which the fraction of sampled problems that
are ready rises steadily, reaching about 70\% by step 200. The critic's error $\varepsilon$
decrease steadily over training.

\subsection{\methodshort{} variants}
\label{sec:variants}

We present three variants of the AC2 algorithm that also 
perform well: (1) training with a highly stale replay buffer,
(2) training without auditing, and (3)
setting group size on ready problems to 1. Results 
are shown in \Cref{fig:component-ablations}, left.

\textbf{AC2 with stale replay buffer.}
Actor--critic methods~\citep{haarnoja2018soft,fujimoto2018addressing}, such as
Soft Actor-Critic (SAC), can learn from rollouts generated by earlier policies.
We test AC2s tolerance to a more stale initial state distribution 
by collecting 1920 fresh rollouts every 10 steps and setting this to 
be the actor replay buffer for those steps
(instead of 192 rollouts per step and a buffer with size 256). The mean 
sore reaches 17.90\% (\Cref{fig:component-ablations}, \textcolor[HTML]{C75D99}{pink line}) and appears to plateau there,
but reaches this value far faster than GRPO.
We conclude that AC2 can learn astale state distribution.

\textbf{AC2 without auditing.}
We try removing auditing from critic 
ready problems 
(\Cref{fig:component-ablations}, \textcolor[HTML]{5F8580}{teal line}).
The run shows slightly larger instability than AC2 run, but it still 
reaches a peak mean of 18.66\% at step 160, outperforming the GRPO baseline. 
We still recommend using auditing as it reaches 
a higher peak reward and allows for tracking the health of 
the critic during training more closely.

\textbf{AC2 with group size 1.}
We next consider removing grouping as well as auditing, giving AC2 w/o Group \& Audit.
For each sampled ready problem, this variant samples 16 separate prefixes
and generates one action chunk $c$ per prefix.
Each continuation uses the advantage estimate
$\hat{A} = v - V^\pi_\theta(s)$, with $v = V^\pi_\theta(s \cdot c)$
or $v = r$ if the action chunk ends the trajectory.
The critic is trained on the same single endpoint, fitting $V^\pi_\theta(s)$ to
$v$ as per the Bellman equation.
We retain full-length groups on unready problems to train the critic using
average terminal rewards.
This run performs similarly to AC2 
(\Cref{fig:component-ablations}, \textcolor[HTML]{6E4C86}{purple line}).

\subsection{Ablations that hurt performance}
\label{sec:component-ablations}

\begin{figure}[t]
\centering
\includegraphics[width=\textwidth]{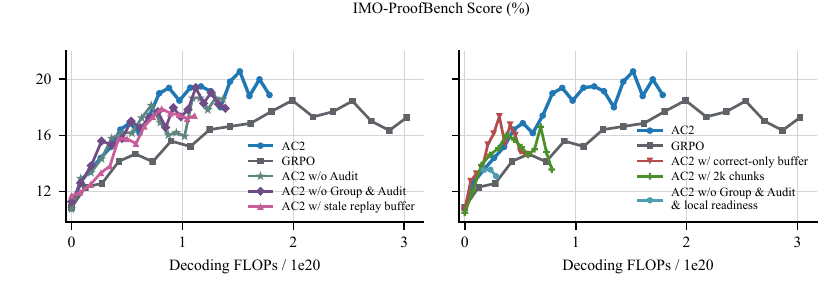}
\caption{\textbf{Component ablations evaluated by mean score versus Decoding FLOPs.}
Left: variants that perform comparably to \textcolor[HTML]{1F77B4}{AC2}.
\textcolor[HTML]{5F8580}{AC2 w/o Audit} removes full-length audits with little effect on performance.
\textcolor[HTML]{6E4C86}{AC2 w/o Group \& Audit} uses one short continuation per ready prefix and the
starting-prefix value as its baseline, achieving comparable scores.
\textcolor[HTML]{C75D99}{AC2 w/ stale replay buffer} refreshes its replay buffer only every ten steps and
continues to improve.
Right: variants that fall behind.
With \textcolor[HTML]{4D9DB3}{AC2 w/o Group \& Audit \& local readiness}, learning stalls and scores decline;
this run branches off \textcolor[HTML]{1F77B4}{AC2} at the checkpoint of step 20.
\textcolor[HTML]{B44C4C}{AC2 w/ correct-only buffer} retains trajectories receiving at least six of
seven judge points in the replay buffer, following~\citet{setlur2026reuse}, and worsens later in training.
\textcolor[HTML]{4A8F29}{AC2 w/ 2k chunks} uses 2,000-token chunks and falls
behind after step 50.
The corresponding training-step plots are in Appendix \Cref{fig:component-ablation-steps}.
}
\label{fig:component-ablations}
\end{figure}

We now ablate the other components of \methodshort to show their necessity. We find 
that the use of the local readiness criterion and
10k long action chunks (as opposed to smaller ones) are essential. Intuitively, local readiness ensures the critic is accurate when 
we use it, and larger action chunks are easier for a the critic to judge, 
and thus make it more accurate, while 
still allowing for finer grained 
credit attribution than rolling out 
the full trajectory. Finally, 
we consider only storing correct trajectories in the actor 
replay buffer like \citet{setlur2026reuse}, but find this  hurts performance.

\textbf{Smaller chunk size hurts performance.}
We test AC2 
with chunk size of 2,000 as opposed to 10,000 used in the main run.
The two runs track each other through step 50, where AC2 w/ 2k chunks
reaches 16.03\%, compared with 16.41\% for AC2's 10,000-token chunks.
Beyond that step the shorter chunks stop improving. AC2 w/ 2k chunks peaks at
16.62\% at step 100 and falls to 13.57\% at step 120
(\Cref{fig:component-ablations}, \textcolor[HTML]{4A8F29}{green line}).

\textbf{AC2 w/ correct-only buffer improves early but worsens later.}
Motivated by \citet{setlur2026reuse}, we keep only trajectories receiving at
least six of seven training-judge points in the replay buffer.
This run reaches 17.35\% at step 60, compared with 16.88\% for AC2.
However, this early advantage does not persist. At step 110, its score falls
to 14.67\%
(\Cref{fig:component-ablations}, \textcolor[HTML]{B44C4C}{red line}).

\textbf{Without local readiness.}
We now consider ablating the use of a local readiness threshold. Specifically, as soon as the critic MAE 
drops below the global $\tau_\text{global}$ threshold, we 
consider all problems critic ready. For this ablation 
we run without group and audit.
Specifically we branch off the AC2
run at step 20 (the  \textcolor[HTML]{1F77B4}{blue line} in
\Cref{fig:component-ablations}. With local readiness disabled, 
mean score
falls from 13.56\% at step 30 to 13.10\% at step 40,
while the run without group and audit but with local readiness reaches 15.33\% at step 40
(\Cref{fig:component-ablations}, \textcolor[HTML]{4D9DB3}{cyan line} versus \textcolor[HTML]{6E4C86}{purple line}).
At that step, only 18.23\% of sampled problems are ready with local readiness,
compared with 100\% (by definition) without it
(Appendix \Cref{fig:additional-diagnostics}).

\begin{figure}[t]
\centering
\includegraphics[width=\textwidth]{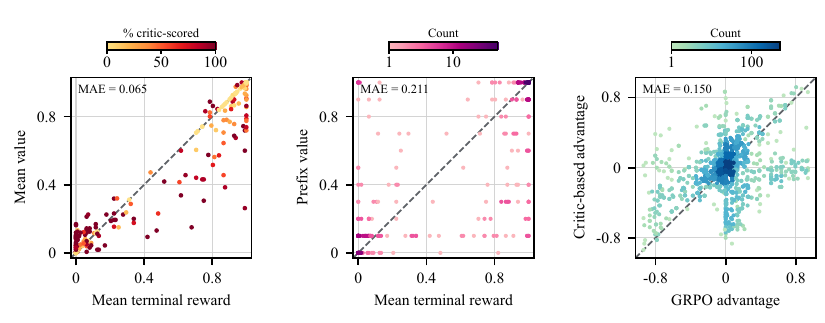}
\caption{\textbf{Group-mean values are more accurate than prefix predictions, and critic-based advantages correlate with GRPO advantages.}
At step 80, the left panel compares $\text{mean}_i(v_i)$ with $\text{mean}_i(r_i)$ for
256 prefixes; colour indicates the percentage of continuations scored by the critic.
The middle panel compares the prefix value prediction $V^\pi_\theta(s)$ with
$\text{mean}_i(r_i)$ for the same prefixes.
The right panel compares $\hat A_i$ with $r_i-\text{mean}_j(r_j)$ for 2,640 responses in
165 groups containing at least one critic prediction.
Details are deffered to \Cref{app:q-probes}.}
\label{fig:critic-diagnostics}
\end{figure}

\subsection{Critic Diagnostics}
\label{sec:q-diagnostics}

\textbf{Group-mean values are more accurate than the prefix value.}
At step 80, we sample 256 ready training problems, cut one prefix $s$ from each
problem's latest stored trajectory, and generate $g=16$ full continuations.
We reconstruct the endpoint value $v_i$ using the critic prediction after 10,000 new tokens when
available and the terminal reward $r_i$ otherwise
(see \Cref{app:q-probes} for details).
Against the mean terminal reward $\text{mean}_i(r_i)$, the group-mean value
$\text{mean}_i(v_i)$ has mean absolute error (MAE) 0.065, compared with 0.211
for the prefix value $V^\pi_\theta(s)$
(\Cref{fig:critic-diagnostics}, left and middle).
This justifies using the group-mean value as supervision for the prefix value.

\textbf{Critic-based advantages are positively correlated with Prefix GRPO advantages.}
Although the prefix value has a larger MAE, the critic-based advantages
$\hat A_i=v_i-\text{mean}_j(v_j)$ have MAE 0.150 relative to the prefix GRPO advantages
$r_i-\text{mean}_j(r_j)$ across 2,640 responses in 165 groups containing at least
one critic prediction (\Cref{fig:critic-diagnostics}, right).
Their Pearson correlation is 0.388.
As AC2 outperforms Prefix GRPO (\Cref{sec:main-results}), Prefix GRPO
advantages should not be treated as ground truth. 

\textbf{A reference solution in the prompt makes the critic more accurate.}
We take the main run's checkpoints (the \textcolor[HTML]{1F77B4}{blue} line in
\Cref{fig:method-overview}) at steps 40, 80, 120 and 160. For each, we collect the
prefixes probed over the next 20 steps whose problem had a reference solution and whose
target is a mean of terminal rewards, about 1,500 per checkpoint. The critic scores each
prefix twice, with and without the reference in its prompt. With the reference, MAE is
lower at every checkpoint (\Cref{fig:main-results}, fourth panel).

\section{Related Work}
\label{sec:related-works}
\label{app:extended-related-work}

\methodshort is an actor-critic method as it learns 
a value model and policy, improving the policy using 
the value model \citep{mnih2016asynchronous, fujimoto2018addressing,
haarnoja2018soft,konda1999actor}. Some readers will have noticed that our method 
name is very similar to
A3C \citep{mnih2016asynchronous}, and this reflects the similarity 
in the methods themselves (although the application settings 
are vastly different). A3C 
estimates the advantage from an $n$-step return with $n$ far smaller than the
episode horizon, using the critic at step $n$. It then uses this 
advantage to improve the actor through a policy gradient update. 
Our use of the GAE $\lambda=0$ advantage estimator on critic 
ready problems to get the advantage of a \emph{single} action chunk 
without rolling out to the end of the trajectory is similar 
in nature. 

Despite their success in continuous control, actor-critic 
methods have not been widely adopted for LLM RLVR. 
Instead, group-relative methods dominate
(specifically GRPO \citep{shao2024deepseekmath,guo2025deepseek} and its descendants
\citep{yu2025dapo, chen2025minimax, khatri2026art,liu2024understanding}). These methods 
sample a group of complete responses per problem and use the group mean of the terminal rewards as a
Monte-Carlo baseline. In the language of \Cref{sec:prelim} this is the
$\lambda = 1$ estimator with the baseline evaluated at the root.
This \emph{requires} every rollout to run to termination, and every token 
of a response receives the same advantage.

In response to the latter concern,
a recent wave of work reintroduces a learned value model to RLVR, but still require calculating 
terminal rewards at each step. Concurrent Le Critique \citep{venkatraman2026critique}, EVPO
\citep{pan2026evpo}, BPCO \citep{qi2026best}, POISE \citep{choi2026your}, JustRL 2 \citep{justrl2_2026},
$V_{0.5}$ \citep{zhang2026v_}, GenAC \citep{shan2026bringing}, and VAPO \citep{yue2025vapo} 
each train a critic alongside the policy but  
advantage estimation still depends on terminal
rewards.
One can also solve the uniform advantage problem without a learnt value model, 
by estimating intermediate values with additional Monte-Carlo rollouts to terminal 
reward. This is the approach taken by
VinePPO \citep{kazemnejad2024vineppo} and SPO
\citep{guo2025spo}, however, this requires generating a large number of tokens for each
advantage calculated. As the value estimate (whether from a value model or 
additional rollouts) is used (primarily) for baseline, all these 
methods sit at or near $\lambda=1$ in \Cref{eq:gae}.
Note any $\lambda > 0$ needs to observe the terminal reward, and 
to our knowledge \methodshort is the first method to date to 
use $\lambda=0$ for LLM RLVR.

Using the value model so aggressively \emph{requires} an 
accurate critic. We achieve this through (1) local 
critic readiness, (2) providing the critic with privileged information (a correct 
solution) if we have it, and (3) action chunking, so the critic judges the state
after a meaningful chunk of reasoning rather than after every token. All three have
some 
precedent in the literature but 
have not been combined.
Conditioning the critic on
privileged information is also done by Le Critique and BPCO
\citep{venkatraman2026critique, qi2026best}. Parameterising the critic as the
language model itself, rather than a scalar head, follows GenAC
\citep{shan2026bringing}. Deciding \emph{when} to trust the critic has been posed
globally, by switching or interpolating baselines on a reliability signal
\citep{zhang2026v_, venkatraman2026critique}
and per-problem by 
\citet{pan2026evpo}, although
their method requires terminal 
rollouts on every problem.
Action chunking \citep{zhao2023learning}, as an intermediate unit of credit assignment between individual
tokens and complete responses, has been explored in reinforcement learning more
broadly by \citet{li2025reinforcement}, and for LLMs by SPO \citep{guo2025spo} and
VinePPO \citep{kazemnejad2024vineppo}, which estimate segment values with additional
Monte-Carlo rollouts rather than a learned critic.

\section{Conclusion}
\label{sec:discussion}

We forward the idea that critics in LLM RL can 
be used more aggressively than prior works.
One example of this is abandoning the requirement 
of collecting a terminal reward, instead using 
a GAE advantage with $\lambda =0 $. Removing 
this requirement opens up the design 
space of LLM RL algorithms, with 
\methodshort being an example of such a new algorithm.
\methodshort also
leverages the special property that you can reset the states during 
LLM RL by sampling action chunks on top of prefixes from 
a replay buffer. One can imagine more exotic 
combinations of a critic and state resetting. For example, 
you could conduct beam search over action chunks using 
an AC2 trained model, or even alpha-go style MCTS tree 
search \citep{silver2016mastering,liu2023don}. 

One of the key components that makes \methodshort work, 
local readiness, is also its largest limitation. In particular, 
local readiness requires epoching the training data as we require  complete rollouts 
 to get a ``ground truth'' 
estimate of the value. In the non data constrained setting, 
this epoching is not realistic. We leave adapting AC2, and more 
generally the local readiness criteria, to the single epoch 
regime as future work.

\section*{Acknowledgments}

KW thanks the support of a Stanford Graduate Fellowship. 
LB thanks the support of a Stanford Graduate 
and Vitalik Buterin Fellowship.
TM thanks the support of NSF 2522743. This work does not necessarily reflect the position or policy of the
government and no official endorsement should be inferred. We thank Google TPU Research Cloud 
and Stanford Marlowe \citep{marlowe2025} for computing resources.

We thank Xingyu Dang, Lars Ankile and Tanishq Kumar 
for helpful feedback throughout the 
project. We thank Neil Band and Caroline Choi, 
for feedback on an early draft of this work. 
We thank Marka Ellertson 
for help with manuscript writing.

\newpage

\ifarxiv\else
\newpage
\subsection*{AI use statement}
In this work, we used generative AI tools for implementing methods
and interpreting results.
We have not used generative AI tools for all other tasks 
listed on the required disclosure.
Additionally, we used generative AI tools for 
    creating figures, initial drafting of the manuscript, summarizing 
    pre-existing literature .
We have reviewed all AI-assisted work. 
LLM-generated code was verified and tested for correctness 
at the thoroughness of human authored code.
We take responsibility for the final content of this work,
including text, claims or artifacts produced with the aid of generative AI.

\subsection*{Reproducibility statement}

\Cref{sec:method} describes our algorithm, and
\Cref{sec:results} specifies the datasets and evaluation
protocol. \Cref{app:implementation} and
\Cref{tab:training-settings} provide implementation details
and hyperparameters; \Cref{app:compute} documents our
compute accounting. We provide an anonymized supplementary
archive containing the training code, experiment configurations,
data-preparation scripts, and reproduction instructions. The
archive also includes recorded measurements and plotting scripts
for the performance and ablation results.
\fi

\bibliography{iclr2027_conference}

\begin{thebibliography}{35}
\providecommand{\natexlab}[1]{#1}
\providecommand{\url}[1]{\texttt{#1}}
\expandafter\ifx\csname urlstyle\endcsname\relax
  \providecommand{\doi}[1]{doi: #1}\else
  \providecommand{\doi}{doi: \begingroup \urlstyle{rm}\Url}\fi

\bibitem[Chen et~al.(2025)Chen, Li, Gong, Jiang, Fei, Yang, Shan, Yu, Wang,
  Zhu, et~al.]{chen2025minimax}
Aili Chen, Aonian Li, Bangwei Gong, Binyang Jiang, Bo~Fei, Bo~Yang, Boji Shan,
  Changqing Yu, Chao Wang, Cheng Zhu, et~al.
\newblock Minimax-m1: Scaling test-time compute efficiently with lightning
  attention.
\newblock \emph{arXiv preprint arXiv:2506.13585}, 2025.

\bibitem[Choi et~al.(2026)Choi, Lim, Ahn, Oh, Shim, and Jo]{choi2026your}
Yunho Choi, Jongwon Lim, Woojin Ahn, Minjae Oh, Jeonghoon Shim, and Yohan Jo.
\newblock Your language model is its own critic: Reinforcement learning with
  value estimation from actor's internal states.
\newblock \emph{arXiv preprint arXiv:2605.07579}, 2026.

\bibitem[DeepSeek-AI et~al.(2026)DeepSeek-AI, Xu, Lin, Xue, Wang, Xu, Wu,
  Zhang, Lin, Dong, Ling, Lu, Zhao, Deng, Hou, Xu, Shao, Ruan, Sun, Dai, Guo,
  Yang, Chen, Li, Ji, Li, Wei, Lin, Yuan, Xia, Dai, Hao, Chen, Cao, Meng, Li,
  Yu, Zhang, Xu, Li, Liang, Zhang, Luo, Wei, Yuan, Zhang, Luo, Chen, Ji, Zhang,
  Ding, Tang, Cao, Gao, Qu, Zeng, Yang, Zhu, Luo, Song, Yu, Huang, Cai, Liang,
  Zhou, Ye, Li, Xu, Hu, Yang, Chen, Yan, Chen, Zhou, Xiang, Yuan, Cheng, Zhou,
  Zhu, Yu, Sun, Ran, Jiang, Qiu, Li, Zheng, Song, Dong, Gao, Guan, Zhou, Huang,
  Yu, Wang, Zhang, Wang, Xia, Zhang, Zhao, Guo, Luo, Ma, Zhu, Wang, Cai, Zhang,
  Chen, Di, Xu, Mei, Wang, Zhang, Zhang, Tang, Li, Zhou, Han, Wang, Huang,
  Wang, Cong, Wang, Zhang, Wang, Zhu, Li, Chen, Du, Jiang, Tian, Xu, Lu, Xu,
  Ge, Zhang, Pan, Wang, Chen, Yin, Xu, Shen, Zhang, Chen, Liu, Lu, Sun, Zhou,
  Chen, Cai, Nie, Wu, Chen, Hu, Liu, Hu, Ma, Wang, Yu, Zhou, Pan, Yu, Zhou, Ni,
  Yun, Jin, Pei, Ye, Lin, Ji, Cui, Yue, Yu, Wang, Zhang, Xiao, Zeng, An, Zhao,
  Liu, Liang, Pang, Luo, Yao, Gao, Yang, Huang, Hou, Zhang, Ma, Gao, He, Wang,
  Wang, Bi, Liu, Wang, Chen, Zhang, Nie, Sun, Wang, Cheng, Liu, Xie, Liu, Liu,
  Yu, Li, Yang, Zhang, Chen, Wang, Su, Chen, Lin, Fu, Yan, Wang, Ma, Luo,
  Zhang, Xu, Ma, Huang, Li, Li, Xu, Zhao, Sun, Wang, Qian, Shao, Yu, Zhang,
  Ding, Shi, Wu, Xiong, Ma, He, Tang, Zhou, Luo, Zhong, Piao, Wang, Zhang,
  Chen, Tan, Wei, Ma, Liu, Yang, Guo, Wu, Wu, Li, Cheng, Ou, Xu, Li, Wang,
  Yang, Xu, Wu, Meng, Zou, Zha, Xiong, Chen, Lin, Cao, Wang, Zhang, Yan, Lin,
  Gu, Luo, You, Liu, Zhou, Zhou, Huang, Wu, Wang, Zhao, Ren, Zhang, Sha, Fu,
  Ju, Xu, Xie, Zhang, Gao, Hao, Gou, Ma, Yan, Shao, Huang, Chen, Wu, Ren, Wu,
  Li, Zhang, Xu, Wang, Qu, Gu, Zhu, Li, Zhang, Xie, Gao, Wan, Pan, and
  Yao]{deepseekai2026deepseekv4highlyefficientmilliontoken}
DeepSeek-AI, Anyi Xu, Bangcai Lin, Bing Xue, Bingxuan Wang, Bingzheng Xu,
  Bochao Wu, Bowei Zhang, Chaofan Lin, Chen Dong, Chenchen Ling, Chengda Lu,
  Chenggang Zhao, Chengqi Deng, Chengyu Hou, Chenhao Xu, Chenze Shao, Chong
  Ruan, Conner Sun, Damai Dai, Daya Guo, Dejian Yang, Deli Chen, Donghao Li,
  Dongjie Ji, Erhang Li, Fang Wei, Fangyun Lin, Fangzhou Yuan, Feiyu Xia,
  Fucong Dai, Guangbo Hao, Guanting Chen, Guoai Cao, Guolai Meng, Guowei Li,
  Han Yu, Han Zhang, Hanwei Xu, Hao Li, Haofen Liang, Haoling Zhang, Haoming
  Luo, Haoran Wei, Haotian Yuan, Haowei Zhang, Haowen Luo, Haoyu Chen, Haozhe
  Ji, Hengqing Zhang, Honghui Ding, Hongxuan Tang, Huanqi Cao, Huazuo Gao, Hui
  Qu, Hui Zeng, J~Yang, JQ~Zhu, Jia Luo, Jia Song, Jia Yu, Jialiang Huang,
  Jialu Cai, Jian Liang, Jiangting Zhou, Jiasheng Ye, Jiashi Li, Jiaxin Xu,
  Jiewen Hu, Jieyu Yang, Jin Chen, Jin Yan, Jingchang Chen, Jingli Zhou,
  Jingting Xiang, Jingyang Yuan, Jingyuan Cheng, Jingzi Zhou, Jinhua Zhu,
  Jiping Yu, Joseph Sun, Jun Ran, Junguang Jiang, Junjie Qiu, Junlong Li,
  Junmin Zheng, Junxiao Song, Kai Dong, Kaige Gao, Kang Guan, Kexing Zhou,
  Kezhao Huang, Kuai Yu, Lean Wang, Lecong Zhang, Lei Wang, Leyi Xia, Li~Zhang,
  Liang Zhao, Lihua Guo, Lingxiao Luo, Linwang Ma, Linyan Zhu, Litong Wang,
  Liyu Cai, Liyue Zhang, Longhao Chen, MS~Di, MY~Xu, Max Mei, Miaojun Wang,
  Mingchuan Zhang, Minghua Zhang, Minghui Tang, Mingming Li, Mingxu Zhou,
  Minmin Han, Ning Wang, Panpan Huang, Panpan Wang, Peixin Cong, Peiyi Wang,
  Peng Zhang, Qiancheng Wang, Qihao Zhu, Qingyang Li, Qinyu Chen, Qiushi Du,
  Qiwei Jiang, Rui Tian, Ruifan Xu, Ruijie Lu, Ruiling Xu, Ruiqi Ge, Ruisong
  Zhang, Ruizhe Pan, Runji Wang, Runqian Chen, Runqiu Yin, Runxin Xu, Ruomeng
  Shen, Ruoyu Zhang, Ruyi Chen, SH~Liu, Shanghao Lu, Shangmian Sun, Shangyan
  Zhou, Shanhuang Chen, Shaofei Cai, Shaoheng Nie, Shaoqing Wu, Shaoyuan Chen,
  Shengding Hu, Shengyu Liu, Shiqiang Hu, Shirong Ma, Shiyu Wang, Shuiping Yu,
  Shunfeng Zhou, Shuting Pan, Shuying Yu, Songyang Zhou, Tao Ni, Tao Yun, Tian
  Jin, Tian Pei, Tian Ye, Tianle Lin, Tianran Ji, Tianyi Cui, Tianyuan Yue,
  Tingting Yu, Tun Wang, W~Zhang, WL~Xiao, Wangding Zeng, Wei An, Weilin Zhao,
  Wen Liu, Wenfeng Liang, Wenjie Pang, Wenjing Luo, Wenjing Yao, Wenjun Gao,
  Wenkai Yang, Wenlve Huang, Wenqing Hou, Wentao Zhang, Wenting Ma, Xi~Gao,
  Xiang He, Xiangwen Wang, Xianzu Wang, Xiao Bi, Xiaodong Liu, Xiaohan Wang,
  Xiaokang Chen, Xiaokang Zhang, Xiaotao Nie, Xiaowen Sun, Xiaoxiang Wang, Xin
  Cheng, Xin Liu, Xin Xie, Xingchao Liu, Xingchen Liu, Xingkai Yu, Xingyou Li,
  Xinyu Yang, Xinyu Zhang, Xu~Chen, Xuanyu Wang, Xuecheng Su, Xueyin Chen,
  Xuheng Lin, Xuwei Fu, YC~Yan, YQ~Wang, YW~Ma, Yanfeng Luo, Yang Zhang,
  Yanhong Xu, Yanru Ma, Yanwen Huang, Yao Li, Yao Li, Yao Xu, Yao Zhao, Yaofeng
  Sun, Yaohui Wang, Yi~Qian, Yi~Shao, Yi~Yu, Yichao Zhang, Yifan Ding, Yifan
  Shi, Yijia Wu, Yiliang Xiong, Yiling Ma, Ying He, Ying Tang, Ying Zhou,
  Yingjia Luo, Yinmin Zhong, Yishi Piao, Yisong Wang, Yixiang Zhang, Yixiao
  Chen, Yixuan Tan, Yixuan Wei, Yiyang Ma, Yiyuan Liu, Yonglun Yang, Yongqiang
  Guo, Yongtong Wu, Yu~Wu, YuKun Li, Yuan Cheng, Yuan Ou, Yuanfan Xu, Yuanhao
  Li, Yuduan Wang, Yuehan Yang, Yuer Xu, Yuhan Wu, Yuhao Meng, Yuheng Zou,
  Yukun Zha, Yunfan Xiong, Yupeng Chen, Yuping Lin, Yuqian Cao, Yuqian Wang,
  Yushun Zhang, Yuting Yan, Yutong Lin, Yuxian Gu, Yuxiang Luo, Yuxiang You,
  Yuxuan Liu, Yuxuan Zhou, Yuyang Zhou, Yuzhen Huang, ZF~Wu, Zehao Wang, Zehua
  Zhao, Zehui Ren, Zekai Zhang, Zhangli Sha, Zhe Fu, Zhe Ju, Zhean Xu, Zhenda
  Xie, Zhengyan Zhang, Zheren Gao, Zhewen Hao, Zhibin Gou, Zhicheng Ma, Zhigang
  Yan, Zhihong Shao, Zhixian Huang, Zhixuan Chen, Zhiyu Wu, Zhizhou Ren,
  Zhongyu Wu, Zhuoshu Li, Zhuping Zhang, Zian Xu, Zihao Wang, Zihua Qu, Zihui
  Gu, Zijia Zhu, Zilin Li, Zipeng Zhang, Ziwei Xie, Ziyi Gao, Ziyi Wan, Zizheng
  Pan, and Zongqing Yao.
\newblock Deepseek-v4: Towards highly efficient million-token context
  intelligence, 2026.
\newblock URL \url{https://arxiv.org/abs/2606.19348}.

\bibitem[Fujimoto et~al.(2018)Fujimoto, van Hoof, and
  Meger]{fujimoto2018addressing}
Scott Fujimoto, Herke van Hoof, and David Meger.
\newblock Addressing function approximation error in actor-critic methods.
\newblock In \emph{International conference on machine learning}, pp.\
  1587--1596. Pmlr, 2018.

\bibitem[Guo et~al.(2025{\natexlab{a}})Guo, Yang, Zhang, Song, Wang, Zhu, Xu,
  Zhang, Ma, Bi, et~al.]{guo2025deepseek}
Daya Guo, Dejian Yang, Haowei Zhang, Junxiao Song, Peiyi Wang, Qihao Zhu,
  Runxin Xu, Ruoyu Zhang, Shirong Ma, Xiao Bi, et~al.
\newblock Deepseek-r1: Incentivizing reasoning capability in llms via
  reinforcement learning.
\newblock \emph{arXiv preprint arXiv:2501.12948}, 2025{\natexlab{a}}.

\bibitem[Guo et~al.(2025{\natexlab{b}})Guo, Xu, Liu, Ye, and Qiu]{guo2025spo}
Yiran Guo, Lijie Xu, Jie Liu, Dan Ye, and Shuang Qiu.
\newblock Segment policy optimization: Effective segment-level credit
  assignment in {RL} for large language models.
\newblock In \emph{Advances in Neural Information Processing Systems},
  volume~38, 2025{\natexlab{b}}.
\newblock \doi{10.52202/085713-3815}.
\newblock URL
  \url{https://proceedings.neurips.cc/paper_files/paper/2025/hash/a6536243037d1e32c20de85137d478da-Abstract-Conference.html}.

\bibitem[Haarnoja et~al.(2018)Haarnoja, Zhou, Abbeel, and
  Levine]{haarnoja2018soft}
Tuomas Haarnoja, Aurick Zhou, Pieter Abbeel, and Sergey Levine.
\newblock Soft actor-critic: Off-policy maximum entropy deep reinforcement
  learning with a stochastic actor.
\newblock In \emph{International conference on machine learning}, pp.\
  1861--1870. Pmlr, 2018.

\bibitem[Kapfer et~al.(2025)Kapfer, Stine, Narasimhan, Mentzel, and
  Candès]{marlowe2025}
Craig Kapfer, Kurt Stine, Balasubramanian Narasimhan, Christopher Mentzel, and
  Emmanuel Candès.
\newblock Marlowe: Stanford's gpu-based computational instrument, 2025.
\newblock URL \url{https://doi.org/10.5281/zenodo.14751899}.

\bibitem[Kazemnejad et~al.(2025)Kazemnejad, Aghajohari, Portelance, Sordoni,
  Reddy, Courville, and Le~Roux]{kazemnejad2024vineppo}
Amirhossein Kazemnejad, Milad Aghajohari, Eva Portelance, Alessandro Sordoni,
  Siva Reddy, Aaron Courville, and Nicolas Le~Roux.
\newblock {VinePPO}: Refining credit assignment in {RL} training of {LLMs}.
\newblock In \emph{Proceedings of the 42nd International Conference on Machine
  Learning}, volume 267, pp.\  29557--29590. PMLR, 2025.
\newblock URL \url{https://proceedings.mlr.press/v267/kazemnejad25a.html}.

\bibitem[Khatri et~al.(2026)Khatri, Madaan, Tiwari, Bansal, Duvvuri, Zaheer,
  Dhillon, Brandfonbrener, and Agarwal]{khatri2026art}
Devvrit Khatri, Lovish Madaan, Rishabh Tiwari, Rachit Bansal, Venkata Sai
  Surya~Subramanyam Duvvuri, Manzil Zaheer, Inderjit Dhillon, David
  Brandfonbrener, and Rishabh Agarwal.
\newblock The art of scaling reinforcement learning compute for llms.
\newblock In \emph{International Conference on Learning Representations},
  volume 2026, pp.\  72438--72467, 2026.

\bibitem[Konda \& Tsitsiklis(1999)Konda and Tsitsiklis]{konda1999actor}
Vijay Konda and John Tsitsiklis.
\newblock Actor-critic algorithms.
\newblock \emph{Advances in neural information processing systems}, 12, 1999.

\bibitem[Li et~al.(2025)Li, Zhou, and Levine]{li2025reinforcement}
Qiyang Li, Zhiyuan Zhou, and Sergey Levine.
\newblock Reinforcement learning with action chunking.
\newblock \emph{arXiv preprint arXiv:2507.07969}, 2025.
\newblock URL \url{https://arxiv.org/abs/2507.07969}.

\bibitem[Liu et~al.(2023)Liu, Cohen, Pasunuru, Choi, Hajishirzi, and
  Celikyilmaz]{liu2023don}
Jiacheng Liu, Andrew Cohen, Ramakanth Pasunuru, Yejin Choi, Hannaneh
  Hajishirzi, and Asli Celikyilmaz.
\newblock Don't throw away your value model! generating more preferable text
  with value-guided monte-carlo tree search decoding.
\newblock \emph{arXiv preprint arXiv:2309.15028}, 2023.

\bibitem[Liu et~al.(2024)Liu, Chen, Li, Qi, Pang, Du, Lee, and
  Lin]{liu2024understanding}
Zichen Liu, Changyu Chen, Wenjun Li, Penghui Qi, Tianyu Pang, Chao Du, Wee~Sun
  Lee, and Min Lin.
\newblock Understanding r1-zero-like training: A critical perspective, 2025.
\newblock \emph{URL https://arxiv. org/abs/2503.20783}, 1, 2024.

\bibitem[{LM-Provers} et~al.(2026){LM-Provers}, Qu, Setlur, Dekoninck,
  Beeching, Li, Wu, Tunstall, and Kumar]{lmprovers2026qednano}
{LM-Provers}, Yuxiao Qu, Amrith Setlur, Jasper Dekoninck, Edward Beeching, Jia
  Li, Ian Wu, Lewis Tunstall, and Aviral Kumar.
\newblock {QED-Nano}: Teaching a tiny model to prove hard theorems.
\newblock \emph{arXiv preprint arXiv:2604.04898}, 2026.
\newblock URL \url{https://arxiv.org/abs/2604.04898}.

\bibitem[Luong et~al.(2025)Luong, Hwang, Nguyen, Ghiasi, Chervonyi, Seo, Kim,
  Bingham, Lee, Mishra, Zhai, Hu, Michalewski, Kim, Ahn, Bae, Song, Trinh, Le,
  and Jung]{luong2025robust}
Thang Luong, Dawsen Hwang, Hoang~H Nguyen, Golnaz Ghiasi, Yuri Chervonyi, Insuk
  Seo, Junsu Kim, Garrett Bingham, Jonathan Lee, Swaroop Mishra, Alex Zhai,
  Huiyi Hu, Henryk Michalewski, Jimin Kim, Jeonghyun Ahn, Junhwi Bae, Xingyou
  Song, Trieu~Hoang Trinh, Quoc~V Le, and Junehyuk Jung.
\newblock Towards robust mathematical reasoning.
\newblock In \emph{Proceedings of the 2025 Conference on Empirical Methods in
  Natural Language Processing}, pp.\  35418--35442. Association for
  Computational Linguistics, 2025.
\newblock \doi{10.18653/v1/2025.emnlp-main.1794}.
\newblock URL \url{https://aclanthology.org/2025.emnlp-main.1794/}.

\bibitem[Mnih et~al.(2016)Mnih, Badia, Mirza, Graves, Lillicrap, Harley,
  Silver, and Kavukcuoglu]{mnih2016asynchronous}
Volodymyr Mnih, Adria~Puigdomenech Badia, Mehdi Mirza, Alex Graves, Timothy
  Lillicrap, Tim Harley, David Silver, and Koray Kavukcuoglu.
\newblock Asynchronous methods for deep reinforcement learning.
\newblock In \emph{International conference on machine learning}, pp.\
  1928--1937. PmLR, 2016.

\bibitem[Ouyang et~al.(2022)Ouyang, Wu, Jiang, Almeida, Wainwright, Mishkin,
  Zhang, Agarwal, Slama, Ray, Schulman, Hilton, Kelton, Miller, Simens, Askell,
  Welinder, Christiano, Leike, and Lowe]{ouyang2022instructgpt}
Long Ouyang, Jeffrey Wu, Xu~Jiang, Diogo Almeida, Carroll~L. Wainwright, Pamela
  Mishkin, Chong Zhang, Sandhini Agarwal, Katarina Slama, Alex Ray, John
  Schulman, Jacob Hilton, Fraser Kelton, Luke Miller, Maddie Simens, Amanda
  Askell, Peter Welinder, Paul Christiano, Jan Leike, and Ryan Lowe.
\newblock Training language models to follow instructions with human feedback.
\newblock In \emph{Advances in Neural Information Processing Systems
  (NeurIPS)}, 2022.

\bibitem[Pan et~al.(2026{\natexlab{a}})Pan, Liu, Lin, Zhu, Zhang, Dou, Gao,
  Han, Wang, Zheng, et~al.]{pan2026evpo}
Chengjun Pan, Shichun Liu, Jiahang Lin, Dingwei Zhu, Jiazheng Zhang, Shihan
  Dou, Songyang Gao, Zhenhua Han, Binghai Wang, Rui Zheng, et~al.
\newblock Evpo: Explained variance policy optimization for adaptive critic
  utilization in llm post-training.
\newblock \emph{arXiv preprint arXiv:2604.19485}, 2026{\natexlab{a}}.

\bibitem[Pan et~al.(2026{\natexlab{b}})]{justrl2_2026}
Haoxuan Pan et~al.
\newblock Justrl-ii: Scaling small llms to 128k reasoning with a critic.
\newblock
  \url{https://panhaoxuan.notion.site/justrl-ii-scaling-small-llms-to-128k-reasoning-with-a-critic},
  2026{\natexlab{b}}.
\newblock Chinese version:
  \url{https://panhaoxuan.notion.site/justrl-ii-small-llms-to-128k-reasoning-with-a-critic-cn}.

\bibitem[Qi et~al.(2026)Qi, Zhou, and Lee]{qi2026best}
Penghui Qi, Xiangxin Zhou, and Wee~Sun Lee.
\newblock Best practice critic optimization.
\newblock \emph{arXiv preprint arXiv:2608.23566}, 2026.

\bibitem[Schulman et~al.(2017)Schulman, Wolski, Dhariwal, Radford, and
  Klimov]{schulman2017ppo}
John Schulman, Filip Wolski, Prafulla Dhariwal, Alec Radford, and Oleg Klimov.
\newblock Proximal policy optimization algorithms.
\newblock \emph{arXiv preprint arXiv:1707.06347}, 2017.

\bibitem[Schulman et~al.(2018)Schulman, Moritz, Levine, Jordan, and
  Abbeel]{schulman2018highdimensionalcontinuouscontrolusing}
John Schulman, Philipp Moritz, Sergey Levine, Michael Jordan, and Pieter
  Abbeel.
\newblock High-dimensional continuous control using generalized advantage
  estimation, 2018.
\newblock URL \url{https://arxiv.org/abs/1506.02438}.

\bibitem[Setlur et~al.(2026)Setlur, Wang, Cohen, Rashidinejad, and
  Xie]{setlur2026reuse}
Amrith Setlur, Zijian Wang, Andrew Cohen, Paria Rashidinejad, and Sang~Michael
  Xie.
\newblock Reuse your {FLOPs}: Scaling {RL} on hard problems by conditioning on
  very off-policy prefixes.
\newblock \emph{arXiv preprint arXiv:2601.18795}, 2026.
\newblock URL \url{https://arxiv.org/abs/2601.18795}.

\bibitem[Shan et~al.(2026)Shan, Zhong, Wang, and Zhao]{shan2026bringing}
Zikang Shan, Han Zhong, Liwei Wang, and Li~Zhao.
\newblock Bringing value models back: Generative critics for value modeling in
  llm reinforcement learning.
\newblock \emph{arXiv preprint arXiv:2604.10701}, 2026.

\bibitem[Shao et~al.(2024)Shao, Wang, Zhu, Xu, Song, Bi, Zhang, Zhang, Li, Wu,
  and Guo]{shao2024deepseekmath}
Zhihong Shao, Peiyi Wang, Qihao Zhu, Runxin Xu, Junxiao Song, Xiao Bi, Haowei
  Zhang, Mingchuan Zhang, Y.~K. Li, Y.~Wu, and Daya Guo.
\newblock {DeepSeekMath}: Pushing the limits of mathematical reasoning in open
  language models.
\newblock \emph{arXiv preprint arXiv:2402.03300}, 2024.

\bibitem[Silver et~al.(2016)Silver, Huang, Maddison, Guez, Sifre, Van
  Den~Driessche, Schrittwieser, Antonoglou, Panneershelvam, Lanctot,
  et~al.]{silver2016mastering}
David Silver, Aja Huang, Chris~J Maddison, Arthur Guez, Laurent Sifre, George
  Van Den~Driessche, Julian Schrittwieser, Ioannis Antonoglou, Veda
  Panneershelvam, Marc Lanctot, et~al.
\newblock Mastering the game of go with deep neural networks and tree search.
\newblock \emph{nature}, 529\penalty0 (7587):\penalty0 484--489, 2016.

\bibitem[Sutton et~al.(1999)Sutton, McAllester, Singh, and
  Mansour]{sutton1999policy}
Richard~S. Sutton, David McAllester, Satinder Singh, and Yishay Mansour.
\newblock Policy gradient methods for reinforcement learning with function
  approximation.
\newblock In \emph{Advances in Neural Information Processing Systems
  (NeurIPS)}, 1999.

\bibitem[{The Microsoft AI Team}(2026)]{mai_thinking_1}
{The Microsoft AI Team}.
\newblock {MAI-Thinking-1}: Building a hill-climbing machine.
\newblock Technical report, Microsoft AI, 2026.
\newblock URL \url{https://microsoft.ai/pdf/mai-thinking-1.pdf}.

\bibitem[Venkatraman et~al.(2026)Venkatraman, Dinot, and
  Aitchison]{venkatraman2026critique}
Siddarth Venkatraman, Matthieu Dinot, and Laurence Aitchison.
\newblock Le critique: Privileged value functions for llm reinforcement
  learning.
\newblock \emph{arXiv preprint arXiv:2608.16739}, 2026.

\bibitem[Yang et~al.(2025)Yang, Li, Yang, Zhang, Hui, Zheng, Yu, Gao, Huang,
  Lv, et~al.]{yang2025qwen3}
An~Yang, Anfeng Li, Baosong Yang, Beichen Zhang, Binyuan Hui, Bo~Zheng, Bowen
  Yu, Chang Gao, Chengen Huang, Chenxu Lv, et~al.
\newblock Qwen3 technical report.
\newblock \emph{arXiv preprint arXiv:2505.09388}, 2025.

\bibitem[Yu et~al.(2025)Yu, Zhang, Zhu, Yuan, Zuo, Yue, Fan, Liu, Liu, Liu,
  Lin, Lin, Ma, Sheng, Tong, Zhang, Zhang, Zhang, Zhu, Zhu, Chen, Chen, Wang,
  Yu, Dai, Song, Wei, Zhou, Liu, Ma, Zhang, Yan, Qiao, Wu, and
  Wang]{yu2025dapo}
Qiying Yu, Zheng Zhang, Ruofei Zhu, Yufeng Yuan, Xiaochen Zuo, Yu~Yue, Tiantian
  Fan, Gaohong Liu, Lingjun Liu, Xin Liu, Haibin Lin, Zhiqi Lin, Bole Ma,
  Guangming Sheng, Yuxuan Tong, Chi Zhang, Mofan Zhang, Wang Zhang, Hang Zhu,
  Jinhua Zhu, Jiaze Chen, Jiangjie Chen, Chengyi Wang, Hongli Yu, Weinan Dai,
  Yuxuan Song, Xiangpeng Wei, Hao Zhou, Jingjing Liu, Wei-Ying Ma, Ya-Qin
  Zhang, Lin Yan, Mu~Qiao, Yonghui Wu, and Mingxuan Wang.
\newblock {DAPO}: An open-source {LLM} reinforcement learning system at scale.
\newblock \emph{arXiv preprint arXiv:2503.14476}, 2025.

\bibitem[Yue et~al.(2025)Yue, Yuan, Yu, Zuo, Zhu, Xu, Chen, Wang, Fan, Du,
  et~al.]{yue2025vapo}
Yu~Yue, Yufeng Yuan, Qiying Yu, Xiaochen Zuo, Ruofei Zhu, Wenyuan Xu, Jiaze
  Chen, Chengyi Wang, TianTian Fan, Zhengyin Du, et~al.
\newblock Vapo: Efficient and reliable reinforcement learning for advanced
  reasoning tasks.
\newblock \emph{arXiv preprint arXiv:2504.05118}, 2025.

\bibitem[Zhang et~al.(2026)Zhang, Sun, Hao, Gu, Cai, Zhan, and Ye]{zhang2026v_}
Yi-Kai Zhang, Yueqing Sun, Hongyan Hao, Qi~Gu, Xunliang Cai, De-Chuan Zhan, and
  Han-Jia Ye.
\newblock {$V_{0.5}$}: Generalist value model as a prior for sparse {RL}
  rollouts.
\newblock \emph{arXiv preprint arXiv:2603.10848}, 2026.

\bibitem[Zhao et~al.(2023)Zhao, Kumar, Levine, and Finn]{zhao2023learning}
Tony~Z Zhao, Vikash Kumar, Sergey Levine, and Chelsea Finn.
\newblock Learning fine-grained bimanual manipulation with low-cost hardware.
\newblock \emph{arXiv preprint arXiv:2304.13705}, 2023.

\end{thebibliography}
\bibliographystyle{iclr2027_conference}

\appendix
\newpage
\section{Additional Experiments}
\label{app:experiments}

\subsection{GRPO tuning}
\label{app:grpo-tuning}

We test GRPO learning rates 
$1\times10^{-6}$,
$2\times10^{-6}$,
$4\times10^{-6}$,
with results shown 
\Cref{fig:grpo-tuning}. 
$2\times10^{-6}$ is 
better than 
$4\times10^{-6}$ 
and comparable to $1\times10^{-6}$
at step 100, so we select 
$2\times10^{-6}$ as our learning rate 
for experiments.

\lukeappendixfig{\textwidth}{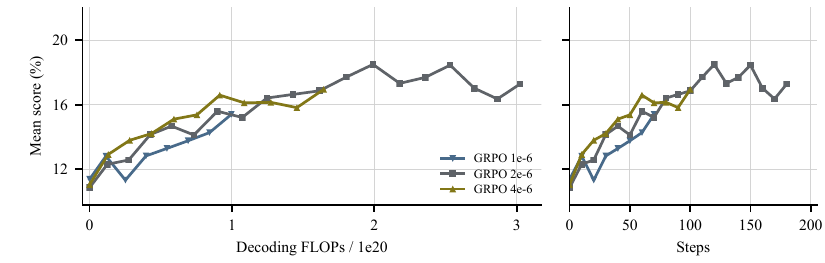}
{\textbf{The observed learning-rate sweep.}
Panels show mean score against Decoding FLOPs and Steps.
}
{fig:grpo-tuning}
\FloatBarrier

\subsection{Additional ablations}
\label{app:additional-ablations}

\lukeappendixfig{\textwidth}{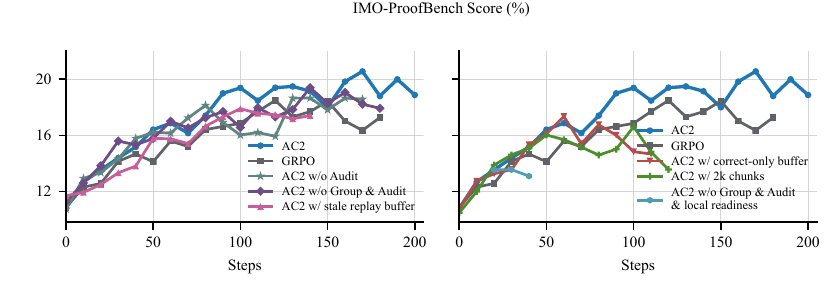}
{\textbf{Component ablations by
steps.}
The panels show the comparisons from \Cref{fig:component-ablations} with
training steps on the horizontal axis. 
}
{fig:component-ablation-steps}

\paragraph{Starting-prefix baseline.}
\Cref{fig:component-ablations} compares AC2 with AC2 w/o Group \& Audit.
Full groups retain 16 continuations per prefix. For each sampled problem assigned to
short generation, AC2 w/o Group \& Audit samples 16 separate prefixes and generates one
continuation from each ($g=1$). This preserves the total number of continuations
per sampled problem.

\paragraph{Chunk size.}
We compare AC2 w/ 2k chunks with AC2, using grouping and auditing in both runs
(\Cref{fig:continuation-budget}). 
AC2 w/ 2k chunks also reduces
the spacing between possible prefix cuts to
2,000 tokens. At step 50, AC2 w/ 2k 
chunks and AC2 reach mean scores
of 16.03\% and 16.41\%, respectively. 
The two runs separate afterwards. 
AC2 w/ 2k chunks peaks at 16.62\% at step
100 and ends at 13.57\% at step 120.
Because this configuration changes the continuation length and the prefix-cut
spacing together, the comparison does not isolate either change.
\lukeappendixfig{\textwidth}{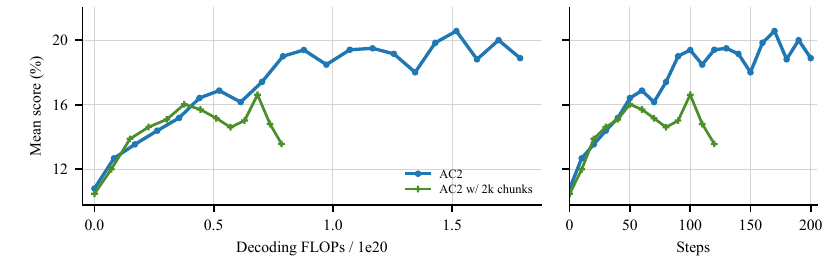}
{\textbf{Chunk-size ablation with grouping and auditing enabled.}
AC2 uses 10,000-token chunks, AC2 w/ 2k chunks uses 2,000-token chunks
and prefix-cut spacing.}
{fig:continuation-budget}

\begin{figure}[H]
\centering
\includegraphics[width=0.5\textwidth]{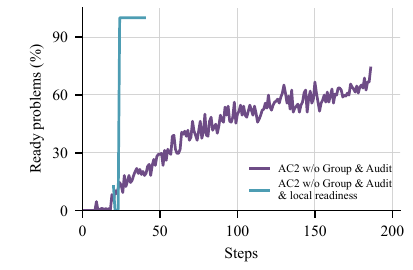}
\caption{\textbf{Removing local readiness.}
AC2 w/o Group \& Audit starts from the base model; the variant that also
removes local readiness branches off AC2 at the checkpoint of step 20.
The plot shows their sampled-problem ready fractions.}
\label{fig:additional-diagnostics}
\end{figure}

\FloatBarrier

\subsection{Additional evaluation metrics}
\label{app:additional-metrics}

We consider 
best-of-16 score as 
an additional metric.
We estimate this with a bootstrap
over 16 sampled validation trajectories,
taking the maximum score of each bootstrapped
sample.
The results are shown in 
\Cref{fig:best-score}.
AC2 continues to be more compute 
efficient in terms of best-of-16,
however AC2 and GRPO plateau at 
roughly the same score.
\lukeappendixfig{\textwidth}{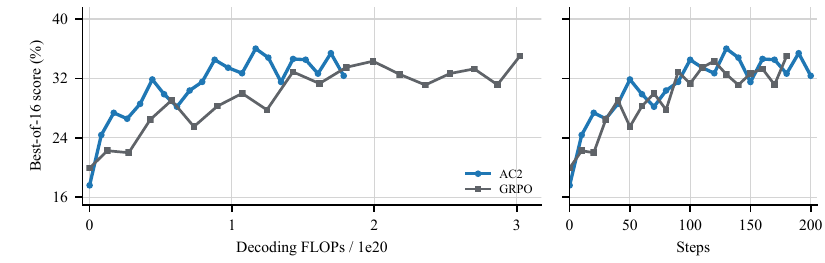}
{\textbf{Best-of-16 scores for AC2 and GRPO.}
}
{fig:best-score}

\subsection{Additional value function diagnostics}
\label{app:q-probes}

We extend the diagnostics in \Cref{sec:q-diagnostics} to steps 40, 57, 80 and 160.
At each checkpoint we select 256 distinct ready problems and their latest trajectories,
cut each trajectory at a random fraction between 5\% and 95\% to give a prefix $s$, and
use the checkpoint's policy to generate 16 full continuations from $s$.
As in the main body, $r_i$ is the terminal reward of the $i$th continuation and $c_i$ is its
first 10k tokens, or fewer if it ends sooner.
Let $v_i = r_i$ if $c_i$ completes the trajectory and $v_i = V^\pi_\theta(s \cdot c_i)$ otherwise,
and let $f$ be the fraction of a group's continuations with $v_i = V^\pi_\theta(s \cdot c_i)$.

\begin{figure}[H]
\centering
\includegraphics[width=0.5\textwidth]{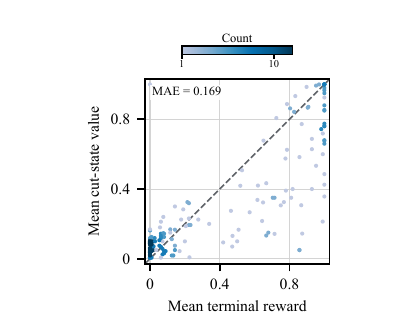}
\caption{\textbf{Group mean of critic values against the group mean of terminal rewards at step 80.}
For the 157 groups with at least two continuations where $v_i = V^\pi_\theta(s \cdot c_i)$, the mean of these $v_i$ (vertical axis) against the mean of the same continuations' $r_i$ (horizontal axis).}
\label{fig:cut-group80}
\end{figure}

At each checkpoint we make the following comparisons.
\begin{itemize}
  \item $V^\pi_\theta(s)$ against $\text{mean}_i(r_i)$ (\Cref{fig:prefix-q-checkpoints}).
  \item $V^\pi_\theta(s)$ against its training target $\text{mean}_i(v_i)$ (\Cref{fig:probe80-reconstruction}, left).
  \item $\text{mean}_i(v_i)$ against $\text{mean}_i(r_i)$ for each group (\Cref{fig:hybrid-early}, and \Cref{fig:probe80-reconstruction}, right).
  \item For groups with at least two continuations where $v_i = V^\pi_\theta(s \cdot c_i)$, the mean of these $v_i$ against the mean of the same continuations' $r_i$ (\Cref{fig:cut-group80}, and \Cref{fig:probe40,fig:probe57,fig:probe160}, left).
  \item $V^\pi_\theta(s \cdot c_i)$ against $r_i$ for each such continuation (\Cref{fig:individual-early,fig:individual-late}).
  \item The critic-based advantage $v_i - \text{mean}_j(v_j)$ against the GRPO advantage $r_i - \text{mean}_j(r_j)$, for groups with $f > 0$ (\Cref{fig:probe40,fig:probe57,fig:probe160}, right).
\end{itemize}
Step-80 versions of the first, third and last comparisons are in \Cref{fig:critic-diagnostics}.
\Cref{tab:probe-summary} lists sample counts and MAEs.
The sampled problems differ between checkpoints, so differences across checkpoints do not
show improvement on a fixed set of problems.

\paragraph{Dependence on $f$.}
A group with $f = 0$ has $v_i = r_i$ for every continuation, so its two group means agree by
construction, and only groups with $f > 0$ test the critic.
\Cref{tab:hybrid-strata} gives the error between $\text{mean}_i(v_i)$ and $\text{mean}_i(r_i)$
separately for $f = 0$, $0 < f < 1$ and $f = 1$ at each checkpoint.

\begin{table}[H]
\centering
\caption{Error between $\text{mean}_i(v_i)$ and $\text{mean}_i(r_i)$ by $f$. Each observation is a group of 16 continuations. Bias is $\text{mean}_i(v_i) - \text{mean}_i(r_i)$, averaged over groups.}
\vspace{5mm}
\label{tab:hybrid-strata}
\begin{tabular}{rlrrr}
\toprule
Step & $f$ & Groups & MAE & Signed bias \\
\midrule
40 & All & 256 & 0.129 & +0.112 \\
40 & $f=0$ & 94 & 0.000 & +0.000 \\
40 & $0<f<1$ & 79 & 0.188 & +0.169 \\
40 & $f=1$ & 83 & 0.218 & +0.183 \\
\midrule
57 & All & 248 & 0.050 & +0.008 \\
57 & $f=0$ & 93 & 0.000 & +0.000 \\
57 & $0<f<1$ & 103 & 0.066 & +0.006 \\
57 & $f=1$ & 52 & 0.107 & +0.026 \\
\midrule
80 & All & 256 & 0.065 & -0.031 \\
80 & $f=0$ & 91 & 0.000 & +0.000 \\
80 & $0<f<1$ & 98 & 0.096 & -0.070 \\
80 & $f=1$ & 67 & 0.106 & -0.018 \\
\midrule
160 & All & 256 & 0.064 & -0.041 \\
160 & $f=0$ & 89 & 0.000 & +0.000 \\
160 & $0<f<1$ & 99 & 0.079 & -0.059 \\
160 & $f=1$ & 68 & 0.126 & -0.066 \\
\bottomrule
\end{tabular}
\end{table}

\lukeappendixfig{0.72\textwidth}{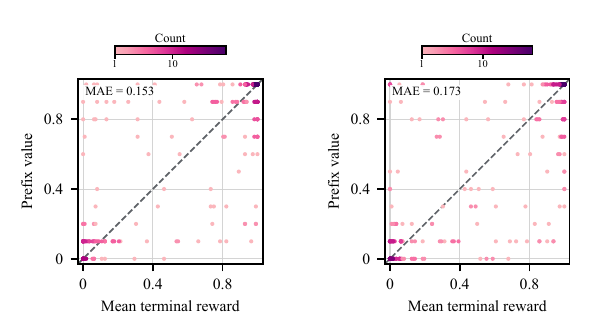}
{\textbf{Critic value at the prefix against the mean terminal reward.} Each point is one prefix $s$, with $V^\pi_\theta(s)$ on the vertical axis and $\text{mean}_i(r_i)$ over its 16 continuations on the horizontal axis. Left: step 57 (248 prefixes). Right: step 160 (256 prefixes). Step 80 is shown in \Cref{fig:critic-diagnostics}, middle.}
{fig:prefix-q-checkpoints}

\lukeappendixfig{0.72\textwidth}{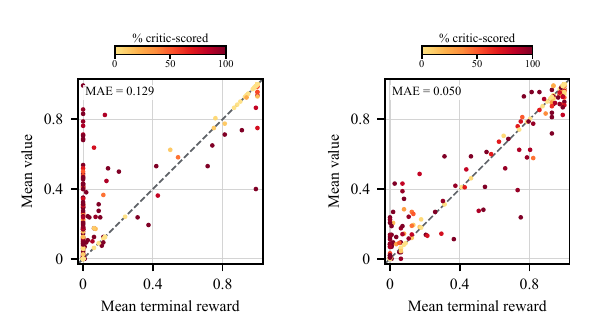}
{\textbf{Group mean of $v_i$ against the group mean of $r_i$.} Each point is one group of 16 continuations, with $\text{mean}_i(v_i)$ on the vertical axis and $\text{mean}_i(r_i)$ on the horizontal axis. Left: step 40. Right: step 57. Colour shows $100f$, the percentage of the group's continuations with $v_i = V^\pi_\theta(s \cdot c_i)$.}
{fig:hybrid-early}

\lukeappendixfig{0.72\textwidth}{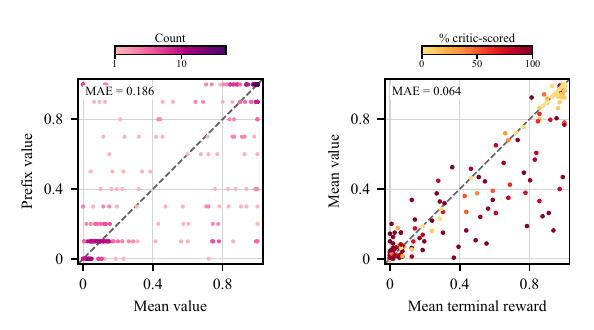}
{\textbf{Critic values and group means at steps 80 and 160.} Left: at step 80, $V^\pi_\theta(s)$ against $\text{mean}_i(v_i)$ for 256 prefixes. Right: at step 160, $\text{mean}_i(v_i)$ against $\text{mean}_i(r_i)$ for 256 groups, with colour showing $100f$.}
{fig:probe80-reconstruction}

\lukeappendixfig{0.72\textwidth}{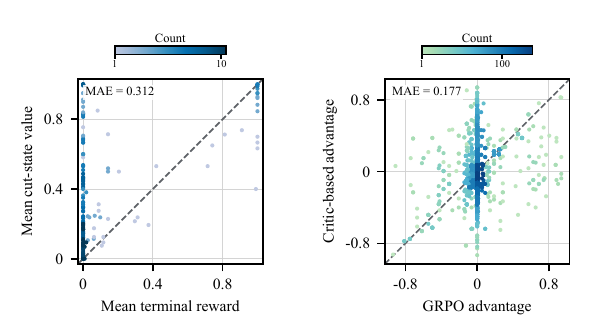}
{\textbf{Value diagnostics at step 40.} Left: for groups with at least two continuations where $v_i = V^\pi_\theta(s \cdot c_i)$, the mean of these $v_i$ against the mean of the same continuations' $r_i$. Right: the critic-based advantage $v_i - \text{mean}_j(v_j)$ against the GRPO advantage $r_i - \text{mean}_j(r_j)$, for groups with $f > 0$.}
{fig:probe40}
\lukeappendixfig{0.72\textwidth}{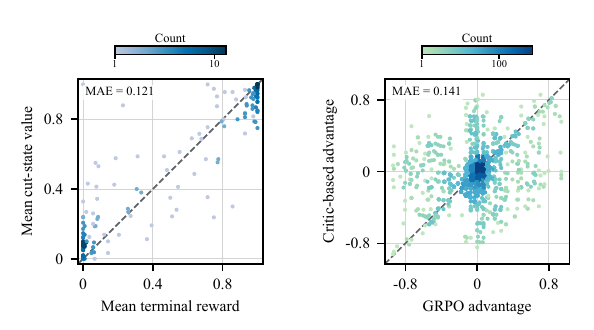}
{\textbf{Value diagnostics at step 57.} Left: for groups with at least two continuations where $v_i = V^\pi_\theta(s \cdot c_i)$, the mean of these $v_i$ against the mean of the same continuations' $r_i$. Right: the critic-based advantage $v_i - \text{mean}_j(v_j)$ against the GRPO advantage $r_i - \text{mean}_j(r_j)$, for groups with $f > 0$.}
{fig:probe57}
\lukeappendixfig{0.72\textwidth}{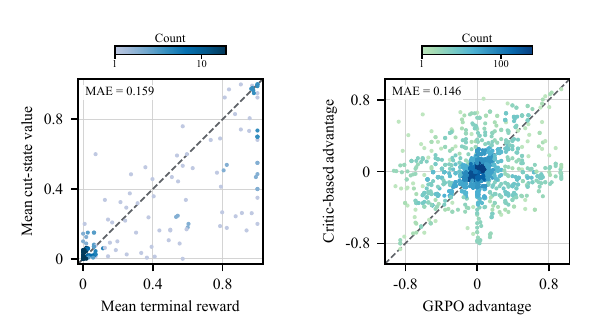}
{\textbf{Value diagnostics at step 160.} Left: for groups with at least two continuations where $v_i = V^\pi_\theta(s \cdot c_i)$, the mean of these $v_i$ against the mean of the same continuations' $r_i$. Right: the critic-based advantage $v_i - \text{mean}_j(v_j)$ against the GRPO advantage $r_i - \text{mean}_j(r_j)$, for groups with $f > 0$.}
{fig:probe160}
\lukeappendixfig{0.72\textwidth}{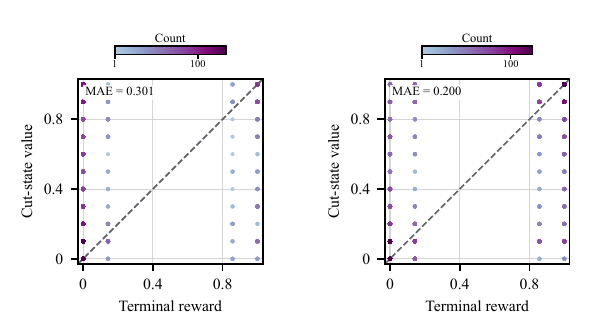}
{\textbf{Critic value at the end of the action chunk against the terminal reward.} Each point is one continuation with $v_i = V^\pi_\theta(s \cdot c_i)$, showing $V^\pi_\theta(s \cdot c_i)$ against $r_i$. Left: step 40. Right: step 57.}
{fig:individual-early}
\lukeappendixfig{0.72\textwidth}{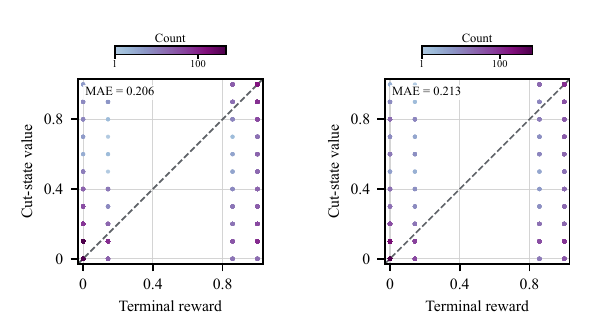}
{\textbf{Critic value at the end of the action chunk against the terminal reward.} Each point is one continuation with $v_i = V^\pi_\theta(s \cdot c_i)$, showing $V^\pi_\theta(s \cdot c_i)$ against $r_i$. Left: step 80. Right: step 160.}
{fig:individual-late}

\begin{table}[t]
\centering
\caption{Sample counts and MAEs for the value diagnostics. Group $n$ counts groups with at least two continuations where $v_i = V^\pi_\theta(s \cdot c_i)$. Centered $n$ and individual $n$ count continuations.}
\label{tab:probe-summary}
\vspace{4mm}
\begin{tabular}{rrrrrr}
\toprule
Step & Group $n$ & MAE & Centered $n$ & MAE & Individual $n$ \\
\midrule
40 & 153 & 0.312 & 2592 & 0.177  & 1919 \\
57 & 144 & 0.121 & 2480 & 0.141  & 1685 \\
80 & 157 & 0.169 & 2640 & 0.150  & 1848 \\
160 & 153 & 0.159 & 2672 & 0.146  & 1892 \\
\bottomrule
\end{tabular}
\end{table}

\FloatBarrier

\newpage
\section{Decoding FLOPs accounting and GPU hours}
\label{app:compute}
\label{app:gpu-hours}

We compute decoding costs from joint prefix and generated-length statistics when
available, and use per-step training-length means otherwise.
AC2 uses joint length statistics throughout. All three GRPO runs, the local-readiness
singleton run from initialization, and AC2 w/ 2k chunks use mean-length
estimates throughout. Prefix GRPO, the no-audit run from initialization,
AC2 w/ correct-only buffer, and AC2 w/ stale replay buffer combine means with retained joint
length records. \Cref{tab:cost-coverage} lists the source and coverage of every
run. These distinctions apply to all FLOPs plots and tables in the paper.

For a policy request, let $p$ count the original prompt and any replay prefix, and
let $g$ count newly generated response tokens. The first output token is sampled
from the prefill computation, leaving $m=\max(g-1,0)$ incremental decoding forwards.
Let $F(p,g)$ denote the decoding FLOP count for this request.
The fixed coefficient $A$ counts attention projections, the gated feed-forward
layers, and vocabulary logits per decoding forward. The coefficient $B$ counts attention matrix
multiplications per context position per decoding forward.
For the Qwen3-4B architecture used here, our matrix-multiplication accounting is
\begin{equation}
F(p,g)=A m+B\left(mp+\frac{m(m+1)}{2}\right),
\qquad A=8{,}044{,}544{,}000,\quad B=589{,}824.
\label{eq:decode-cost}
\end{equation}
The architecture has 36 layers, hidden dimension 2,560, feed-forward dimension 9,728, 32 query heads,
eight key/value heads, head dimension 128, and vocabulary size 151,936. A multiply
and an addition count as two FLOPs. We exclude prefill, validation, value function and judge
calls, training forward/backward computation, elementwise operations, communication.

We get the cumulative decoding FLOPs 
by summing over all actor generated 
tokens, including those to refill 
the buffer $\mathcal{B}$.
Where paired per-request prompt-plus-prefix and generated lengths
are unavailable, let $N$ be the number of completed requests and let $\bar m$ and
$\bar p$ be their empirical mean decoding-forward count and prompt-plus-prefix length.
Let $F_{\mathrm{total}}$ be the sum of their per-request decoding costs.
We estimate this total by substituting the means into the cost formula, giving the estimate $\widehat F$,
\begin{equation}
\widehat F=N\left\{A\bar m+B\left[\bar m\bar p+
\frac{\bar m^2+\bar m}{2}\right]\right\}.
\label{eq:mean-cost}
\end{equation}
For empirical moments defined with denominator $N$, the omitted term is
\begin{equation}
F_{\mathrm{total}}-\widehat F
=NB\left[\operatorname{Cov}(m,p)+\tfrac12\operatorname{Var}(m)\right].
\label{eq:mean-cost-error}
\end{equation}
Because $\mathrm{Cov}(m,p)$ can be negative, $\widehat F$ can over- or underestimate the true cost.
The per-step approximation uses total generated tokens divided by request count and weights
replay-only prefix means by the fraction of replay requests. It assumes each request generates
at least one token and the logged prompt mean represents all requests.
Where complete records exist, the mean-based estimate falls 4.01\% below the exact cost for the
selected GRPO run (steps 161--180) and 5.51\% below for the main \methodshort run.
All GRPO curves use the mean-based estimate and \methodshort uses exact records, so GRPO's
plotted cost is, if anything, slightly lower
than the true value.
\begin{table}[t]
\centering
\caption{Available decoding-cost records. Joint lengths retain per-request prefix/decode moments. Step means yield estimated costs. The last cost-data step can exceed the last evaluation.}
\label{tab:cost-coverage}
\vspace{2mm}
\begin{tabular}{llr}
\toprule
Configuration & Length source & Last cost step \\
\midrule
AC2 & Joint lengths & 202 \\
GRPO, $10^{-6}$ & Step means & 70 \\
GRPO, $2\times10^{-6}$ & Step means & 180 \\
GRPO, $4\times10^{-6}$ & Step means & 100 \\
Prefix GRPO & Means + joint lengths & 120 \\
No audit, branch at 40 & Joint lengths & 123 \\
No audit, from initialization & Means + joint lengths & 173 \\
AC2 w/ correct-only buffer & Means + joint lengths & 117 \\
\shortstack[l]{AC2 w/o Group \& Audit\\(step-50 branch)} & Joint lengths & 117 \\
\shortstack[l]{AC2 w/o Group \& Audit\\(from base model)} & Step means & 186 \\
\shortstack[l]{AC2 w/o Group \& Audit\\\& local readiness (step-20 branch)} & Joint lengths & 40 \\
AC2 w/ 2k chunks & Step means & 128 \\
AC2 w/ stale replay buffer & Means + joint lengths & 146 \\
75k response limit & Joint lengths & 161 \\
\bottomrule
\end{tabular}
\end{table}

\FloatBarrier
\paragraph{Relation to GPU hours.}

We compare Decoding FLOPs with GPU hours for the main \methodshort{} run.
For each training step, let $D$ denote its policy-rollout decoding cost from
\Cref{eq:decode-cost}, summed over requests, and let $H$ denote its GPU
hours: the logged training-step duration in hours multiplied by 32 GPUs.
The timer includes generation, value inference, judging, parameter updates, and
checkpointing, but excludes validation, startup, and failed attempts.
We use 196 complete steps from steps 1--202, excluding six steps resumed from
cached intermediate results because their timers omit earlier computation.

\begin{figure}[H]
\centering
\includegraphics[width=0.62\textwidth]{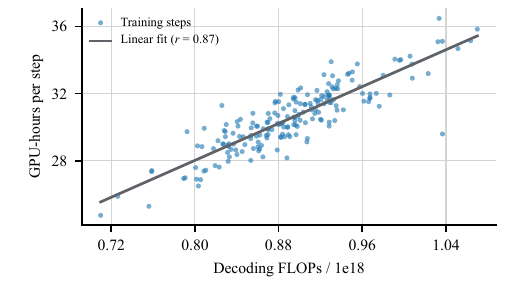}
\caption{\textbf{GPU-hours against Decoding FLOPs per training step.}
Each point is one training step of the main \methodshort{} run. The line is the least-squares fit.}
\label{fig:gpu-hours}
\end{figure}

We fit $H$ as a linear function of $D$ by ordinary least squares with an intercept.
The predicted GPU hours, $\widehat H$, are
$\widehat H=6.11+27.41\,(D/10^{18})$.
The per-step Pearson correlation is $r=0.874$,
(\Cref{fig:gpu-hours}).
This association supports Decoding FLOPs as a proxy for training cost in this run.

\FloatBarrier
\newpage
\section{Training and evaluation details}
\label{app:implementation}

\begin{table}[H]
\centering
\caption{Main AC2 configuration and evaluation settings. Symbols follow
\Cref{sec:method}. The total response budget includes replayed prefixes, with
$b$ counting new tokens. Settings without a symbol in \Cref{sec:method}
are listed by name.}
\vspace{5mm}
\label{tab:training-settings}
\begin{tabular}{@{}p{2.00in}p{0.72in}p{2.42in}@{}}
\toprule
Parameter & Symbol & Value \\
\midrule
Group size & $g$ & 16 continuations per prefix \\
Audit fraction & $\alpha$ & $1/4$; count of ready problems rounded up \\
Replayed prefixes per step & $n_{\text{batch}}$ & 192 \\
Fresh problems per step & $n_{\text{refill}}$ & 192 problems, one rollout each \\
Replay buffer capacity & --- & 256 trajectories \\
Total response budget & --- & 50,000 tokens \\
Action-chunk length & $b$ & 10,000 new tokens \\
Prefix cut grid / maximum fraction & --- & 10,000 tokens / 0.9 \\
\midrule
Critic value grid & --- & $\{0,0.1,\ldots,1\}$ \\
Critic buffer capacity & --- & 1,920 prefix--target pairs \\
Prefix--target pairs per critic update & --- & Up to 768 \\
Valid rewards needed for a critic target & --- & At least 8 \\
Actor learning rate & --- & $2\times10^{-6}$ \\
Initial critic learning rate & --- & $2\sqrt{2}\times10^{-6}$ \\
Critic learning-rate floor & --- & $5\sqrt{2}\times10^{-7}$ \\
Actor minibatches & --- & 1,536 continuations each \\
Critic gradient-norm clip & --- & 0.2 \\
Critic decoding & --- & Greedy, at most four tokens \\
Readiness window & --- & 5 training steps \\
Global readiness threshold & $\tau_{\mathrm{global}}$ & 0.20 \\
Local readiness threshold & $\tau_{\mathrm{local}}$ & 0.18 \\
\midrule
Training temperature / top-$p$ / top-$k$ & --- & 0.8 / 1 / unrestricted \\
Evaluation responses per problem & --- & 16 \\
Evaluation interval & --- & 10 training steps \\
Evaluation temperature / top-$p$ / top-$k$ & --- & 0.8 / 0.95 / 20 \\
Evaluation response limit & --- & 50,000 tokens \\
\bottomrule
\end{tabular}
\end{table}

\paragraph{Replay buffer.}
The replay buffer $\mathcal{B}$, the critic's buffer of prefix--target pairs 
and the
bank of reference solutions all start 
empty. While $\mathcal{B}$ is
empty, the $n_{\text{batch}}$ training prefixes are replaced by fresh problems with empty responses.
Once $\mathcal{B}$ holds trajectories, we form $\mathcal{S}_{\text{batch}}$ by randomly permuting the
distinct problems in $\mathcal{B}$ and cycling through them until $n_{\text{batch}}$ problems are
chosen, then taking one of each problem's stored trajectories uniformly at random.
For a trajectory of $L$ response tokens, the cut position is drawn uniformly from the multiples of
10,000 tokens in $[0, 0.9L]$, where 0 gives an empty prefix. $\mathcal{B}$ holds up to 256
trajectories and replaces the oldest first. Rollouts from $\mathcal{S}_{\text{refill}}$ enter
$\mathcal{B}$ whether or not they are correct. The 50,000-token response budget counts both the
prefix and the newly generated tokens.

\paragraph{Reference solutions.}
Separately from $\mathcal{B}$, we keep a bank holding one reference solution per problem.
A rollout from either $\mathcal{S}_{\text{refill}}$ or $\mathcal{S}_{\text{batch}}$ can supply a
reference if the judge awards it at least six of seven points. For each problem without a
reference, we store the proof from one such rollout, chosen at random. Existing references are
never replaced. Readiness at a step uses the bank as it was before that step's additions, while
that step's critic update can already use the new references.

\paragraph{Critic targets.}
A prefix yields a critic target only if at least 8 of its $g$ continuations have a valid reward.
In the group-size-1 variant, the single continuation must be valid, and a continuation whose
baseline $V^\pi_\theta(s)$ is invalid is left out of both the critic target and the actor loss.
Targets are rounded to the nearest value in $\{0, 0.1, \dots, 1\}$, with ties rounded down.
Each critic update samples up to 768 prefix--target pairs, without replacement, from its buffer of
1,920, skipping pairs that exceed the context limit. A pair uses the reference solution of its
source trajectory when there is one, and otherwise the bank's current reference for its problem,
so an older pair can gain a reference later. As in \Cref{sec:updates}, a pair with a reference is
trained under both prompts at weight $1/2$ each, and a pair without one under the plain prompt
at weight 1. The two prompts are given below. The loss is the negative log-likelihood of the value tokens and the end-of-turn
token, divided by four (the maximum 
number of tokens for a completion). At inference, the critic greedily generates at most four tokens and
we parse the first valid grid value. 

\paragraph{Critic prompt.}
The critic's context is the problem prompt, followed by the partial response and a closing
\texttt{</think>} if the response has none, followed by the instruction below as a new user turn.
The critic's reply is prefilled with an empty thinking section and the answer stem
``\texttt{Q value:~}'', so the critic only generates the number.
When a reference solution is available, the prompt includes it in place of
\texttt{\{reference\_proof\}}.

\noindent\fbox{\parbox{\dimexpr\linewidth-2\fboxsep-2\fboxrule\relax}{\small\raggedright
\textbf{With reference solution}\\[0.4em]
Pause here and estimate how much rubric credit this attempt will earn if you continue it to completion within its remaining token budget: 0 means no credit, 1 means full credit for a complete and correct proof, and values in between mean partial credit. You should refer to a reference correct proof here:\\[0.6em]\{reference\_proof\}\\[0.6em]Don't think; answer immediately with a single line and nothing else:\\[0.6em]Q value: z\\[0.6em]where z is one of 0, 0.1, \ldots, 1. The first characters of your response must be \textasciigrave Q value:\textasciigrave.
}}

\vspace{0.5em}
\noindent\fbox{\parbox{\dimexpr\linewidth-2\fboxsep-2\fboxrule\relax}{\small\raggedright
\textbf{Without reference solution}\\[0.4em]
Pause here and estimate how much rubric credit this attempt will earn if you continue it to completion within its remaining token budget: 0 means no credit, 1 means full credit for a complete and correct proof, and values in between mean partial credit.\\[0.6em]Don't think; answer immediately with a single line and nothing else:\\[0.6em]Q value: z\\[0.6em]where z is one of 0, 0.1, \ldots, 1. The first characters of your response must be \textasciigrave Q value:\textasciigrave.
}}

\paragraph{Readiness and auditing.}
The critic error $\varepsilon$ of \Cref{sec:sampling} compares the critic's prediction before
that step's update with the rounded group target, and an invalid prediction counts as error 1.
Global readiness averages $\varepsilon$ over every prefix sampled in the last five complete steps,
with $\tau_{\text{global}} = 0.20$, and local readiness uses $\tau_{\text{local}} = 0.18$.
Besides the three conditions in \Cref{sec:sampling}, a problem also needs a nonzero critic
prediction at the current or an earlier step. 
In our main AC2 run, due to a bug, we only activate the 
reference proof readiness condition at step 42. As per 
\Cref{fig:main-results} right, which shows reference solutions 
decrease critic MAE, we expect this only hurts the performance 
of our reported AC2 results compared to running with 
this criteria from step 1.
Once ready, a problem stays ready.
Of the $n$ ready problems sampled at a step, $\lceil \alpha n \rceil$ are audited, with
$\alpha = 1/4$, and receive full rollouts. The other ready problems receive action chunks of at
most $b$ new tokens, capped by the remaining response budget, and unready problems always receive
full rollouts. A continuation's endpoint is scored by the critic, $v_i = V^\pi_\theta(s \cdot c_i)$, only if it
reaches $b$ tokens before the response budget runs out and contains no complete proof, that is,
no \texttt{<proof>}\dots\texttt{</proof>} block after the thinking section. Continuations that
finish, exhaust the response budget or contain a complete proof are scored by the judge.

\paragraph{Scoring failures.}
A response with no proof receives reward 0. If the judge request fails or its output cannot be
parsed, the response is left out of critic targets but still enters the actor update with reward 0.
If the critic's value at a continuation's endpoint is invalid, that continuation gets zero weight
in the actor loss, and its $v_i$ is replaced by the mean of the valid $v_j$ in its group, or by 0
if none are valid, before $\text{mean}_j(v_j)$ is computed. A group with no valid values gives no
critic target. %

\paragraph{Optimization.}
Each step applies the first actor minibatch update, one critic update and then the second actor
minibatch update, all to the shared parameters $\theta$ with separate optimizer states.
Each actor minibatch has 1,536 continuations. Advantages are computed on the whole batch before it
is split into minibatches, so the continuations of one group can fall into both. The actor loss is
averaged over generated tokens in each minibatch, with one epoch and no minibatch shuffling.
The critic uses AdamW with betas $(0.9, 0.999)$, epsilon $10^{-8}$ and no weight decay.
Let $\Delta\theta_V$, $\Delta\theta_{\mathrm{PPO},1}$ and $\Delta\theta_{\mathrm{PPO},2}$ be the
parameter changes from the critic update and the first and second actor updates, each measured from
the parameters just before that update. The critic learning rate starts at $2\sqrt{2}\times10^{-6}$
and is halved, down to the floor in \Cref{tab:training-settings}, after two consecutive steps in which
$\lVert\Delta\theta_V\rVert_2 /
(\lVert\Delta\theta_{\mathrm{PPO},1}\rVert_2+\lVert\Delta\theta_{\mathrm{PPO},2}\rVert_2)>\sqrt{2}$.
The actor uses gradient clipping at 0.3 and weight decay 0.01, with no KL penalty or entropy bonus,
and a dual-clip coefficient of 3.

\paragraph{Adaptive entropy control.}
Let $\widehat H$ be the mean entropy of the actor's token distribution, in nats, at the sampling
temperature and before the update. It is averaged over generated tokens in responses with an empty
prefix, namely the rollouts from $\mathcal{S}_{\text{refill}}$ and prefixes cut at position 0.
Continuations of nonempty prefixes are excluded. Given target entropy $H^\star$, step size
$\delta_H$, an offset $k$ bounded by $k_{\min}$ and $k_{\max}$, and base upper clipping parameter
$\epsilon_{\mathrm{high}}^{\mathrm{base}}$, each step updates
\begin{equation}
 \begin{aligned}
 k &\leftarrow \mathrm{clip}\!\left(k+\delta_H\,\mathrm{sign}(H^\star-\widehat H),
 k_{\min},k_{\max}\right),\\
 \epsilon_{\mathrm{high}}&=\epsilon_{\mathrm{high}}^{\mathrm{base}}+k.
 \end{aligned}
 \label{eq:adaptive-clip}
\end{equation}
The update runs once per step before the actor update, so both actor minibatches use the same
$\epsilon_{\mathrm{high}}$, and $\epsilon_{\mathrm{low}}$ stays fixed.
We use $\epsilon_{\mathrm{low}}=0.2$, $\epsilon_{\mathrm{high}}^{\mathrm{base}}=0.28$, $H^\star=0.28$,
$\delta_H=0.02$, $k_{\min}=-0.08$ and $k_{\max}=0.08$, and initialise $k=0.06$.
The offset persists across steps and across resumes from checkpoints.

\paragraph{Evaluation and baseline.}
Evaluation draws 16 responses per problem every ten training steps.
The main comparison uses a 50,000-token evaluation response limit.
GRPO samples 256 fresh problems with 16 responses per problem at each training
step, whereas AC2 trains on the $n_{\text{batch}}=192$ groups in
\Cref{tab:training-settings}. Equal Steps therefore need not imply equal
numbers of responses or equal decoding work.

\clearpage

\end{document}